\documentclass[a4paper,fleqn]{cas-dc}

\usepackage[authoryear,longnamesfirst]{natbib}

\usepackage{graphicx} % Required for inserting images
\usepackage{algorithm}
\usepackage{algorithmic}
\usepackage{float}
\usepackage{mathtools}
\usepackage{caption}
\usepackage{subcaption}
\usepackage{setspace}
\usepackage{amssymb}
\usepackage{latexsym}

\usepackage{bm}
\boldmath

\usepackage{booktabs}
\usepackage{multirow}

\DeclarePairedDelimiter\ceil{\lceil}{\rceil}

\newcommand{\Tc}{\textbf{\textit{T}}^{[c]}}
\newcommand{\tci}{t^{[c]}_{1}}
\newcommand{\TcSet}{\{t^{[c]}_{1}, t^{[c]}_{2}, \ldots, t^{[c]}_{p}\}}
\newcommand{\Tf}{\textbf{\textit{T}}^{[c]}_{filtered}}

\newcommand{\Sc}{\textbf{\textit{S}}^{[c]}}
\newcommand{\sci}{s^{[c]}_{1}}
\newcommand{\ScSet}{\{s^{[c]}_{1}, s^{[c]}_{2}, \ldots, s^{[c]}_{q}\}}
\newcommand{\SBoldD}{\{{\textbf{\textit{S}}^{[c_1]}}, {\textbf{\textit{S}}^{[c_2]}}, \ldots, {\textbf{\textit{S}}^{[c_m]}}\}}

\newcommand{\RBold}{\textbf{\textit{R}}}
\newcommand{\alphaBold}{\mathcal{\alpha}}
\def\tsc#1{\csdef{#1}{\textsc{\lowercase{#1}}\xspace}}
\tsc{WGM}
\tsc{QE}
\begin{document}
\let\WriteBookmarks\relax
\def\floatpagepagefraction{1}
\def\textpagefraction{.001}

% Short title
\shorttitle{CosSIF}    

% Short author
\shortauthors{M. Islam et al.}  

% Main title of the paper
\title [mode = title]{CosSIF: Cosine similarity-based image filtering to overcome low
inter-class variation in synthetic medical image datasets}  

\author[1]{Mominul Islam}[orcid=0009-0001-6409-964X]
\cormark[1] 
\ead{mominul.ivi@gmail.com} 

\author[2]{Hasib Zunair}[orcid=0000-0002-5984-1731]
\ead{hasibzunair@gmail.com} 

\author[1] {Nabeel Mohammed}[orcid=0000-0002-7661-3570]
\ead{nabeel.mohammed@northsouth.edu} 

\address[1]{Department of Electrical and Computer Engineering, North South University, Bashundhara, Dhaka, Bangladesh}
\address[2]{Concordia Institute for Information Systems Engineering, Concordia University, Montreal, QC, Canada}

\cortext[cor1]{Corresponding author} 

% Here goes the abstract
\begin{abstract}
Crafting effective deep learning models for medical image analysis is a complex task, particularly in cases where the medical image dataset lacks significant inter-class variation. This challenge is further aggravated when employing such datasets to generate synthetic images using generative adversarial networks (GANs), as the output of GANs heavily relies on the input data. In this research, we propose a novel filtering algorithm called Cosine Similarity-based Image Filtering (CosSIF). We leverage CosSIF to develop two distinct filtering methods: Filtering Before GAN Training (FBGT) and Filtering After GAN Training (FAGT). FBGT involves the removal of real images that exhibit similarities to images of other classes before utilizing them as the training dataset for a GAN. On the other hand, FAGT focuses on eliminating synthetic images with less discriminative features compared to real images used for training the GAN. Experimental results reveal that employing either the FAGT or FBGT method with modern transformer and convolutional-based networks leads to substantial performance gains in various evaluation metrics. FAGT implementation on the ISIC-2016 dataset surpasses the baseline method in terms of sensitivity by 1.59\% and AUC by 1.88\%. Furthermore, for the HAM10000 dataset, applying FABT outperforms the baseline approach in terms of recall by 13.75\%, and with the sole implementation of FAGT, achieves a maximum accuracy of 94.44\%. Code and implementation details are available at: \url{https://github.com/mominul-ssv/cossif}.
\end{abstract}

% Use if graphical abstract is present
%\begin{graphicalabstract}
%\includegraphics{}
%\end{graphicalabstract}

% Research highlights
% \begin{highlights}
% \item 
% \item 
% \item 
% \end{highlights}

% Keywords
% Each keyword is seperated by \sep
\begin{keywords}
Medical image datasets\sep 
Skin lesion classification\sep
Cosine similarity\sep
Generative adversarial networks\sep
Vision transformer\sep
Swin transformer\sep
ConvNeXt\sep
\end{keywords}

\maketitle

% Main text
\section{Introduction}

\begin{figure*}[t!]
    \centering
    \includegraphics[width=0.88\textwidth]{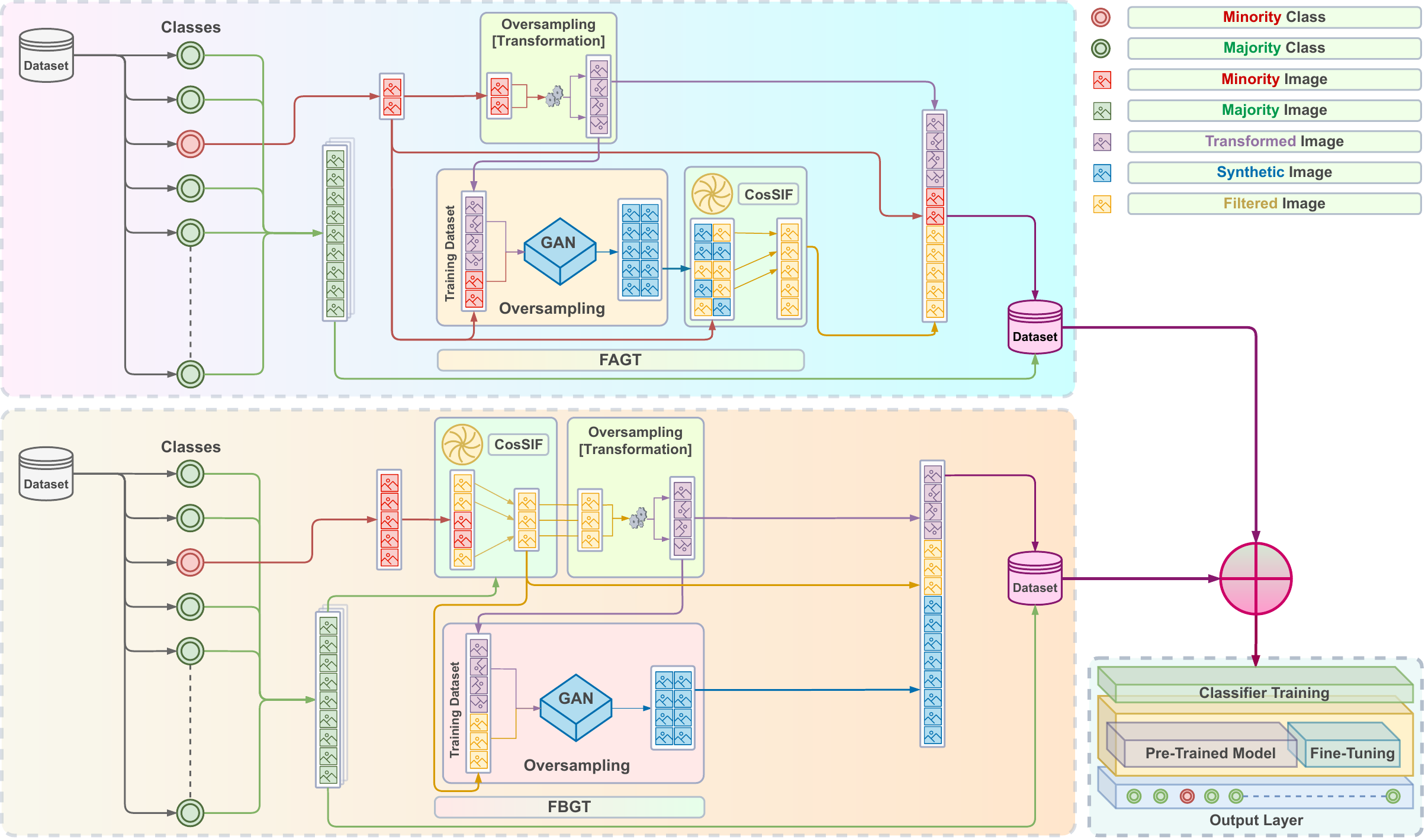}
    \caption{The illustration depicts the pipeline of our research, which involves identifying the minority class from the dataset, oversampling through GAN and transformation techniques, the adoption of our proposed FBGT and FAGT methods to mitigate low inter-class variation by leveraging our novel CosSIF algorithm, and ultimately training classifiers using the augmented dataset. In the case of multiple minority classes, the pipeline is repeated until the classifier training stage. It is recommended to view the illustration in color.}
    \label{fig:pipeline}
\end{figure*}

Medical image analysis is a critical component of modern healthcare, enabling accurate diagnosis, effective treatment, and continuous monitoring of various diseases \cite{bir2020review}. The advent of deep learning has created a new horizon in this field, delivering significant improvements in the early detection and classification of diseases. Numerous studies have highlighted the efficacy of deep learning in medical imaging \cite{adegun2021deep, razzak2018deep}, leading to its widespread adoption in multiple medical domains, including radiology, dermatology, and ophthalmology \cite{mcbee2018deep, lopez2017skin, wang2022artificial}. In radiology, deep learning models have surpassed the diagnostic accuracy of radiologists in detecting breast cancer in mammography images \cite{mckinney2020international}. Similarly, in dermatology, deep learning has demonstrated outstanding performance in identifying skin cancer from dermoscopy images \cite{esteva2017dermatologist}. In ophthalmology, deep learning models have been used to diagnose diabetic retinopathy and age-related macular degeneration from retinal images \cite{gulshan2016development, litjens2017survey}.

One of the major challenges in developing deep learning models for medical image analysis is the limited availability of datasets. This issue is particularly significant in classification tasks, where obtaining a balanced dataset with high inter-class variation is difficult \cite{huynh2022semi}. Inter-class variation refers to the differences in appearance between different classes of images \cite{deng2009imagenet}. The scarcity of a balanced dataset arises when obtaining a sufficient number of images from certain classes is challenging, resulting in an imbalanced distribution of classes. For example, Zech et al. \cite{zech2018variable} reported that training a deep learning model to detect pneumonia in chest radiographs was significantly impacted by the imbalanced nature of the dataset. The authors noted that obtaining a balanced dataset with a sufficient number of images of the positive class was difficult due to the low prevalence of pneumonia in the population. Therefore, low inter-class variation and class imbalance in medical image datasets significantly undermine the applicability of deep learning techniques in medical imaging. 

The use of generative adversarial networks (GANs) has increasingly gained popularity in recent years to address class imbalances in datasets \cite{sampath2021survey}. GANs are generative models that can produce synthetic images that closely resemble real images \cite{karras2017progressive}. Several studies have reported success using GANs to address the class imbalance in various domains, including medical image analysis \cite{kazeminia2020gans}. The conventional approach for training a GAN involves utilizing every image of the minority class in a dataset. \cite{creswell2018generative}. However, when dealing with images that have low inter-class variation, this method can be problematic. This is because the GAN may generate synthetic images that visually resemble images from other classes, resulting in a dataset that is technically balanced but presents challenges for a neural network attempting to distinguish differences between classes. The is largely attributed to the lack of diversity in the synthetic images that makes it difficult for a network to learn discriminative features required for accurate classification.

In response to the challenge of GANs producing visually similar images with less discriminative features when trained on datasets with low inter-class variation, we propose a novel filtering algorithm called Cosine Similarity-based Image Filtering (CosSIF). We utilize CosSIF to introduce two filtering methods: Filtering Before GAN Training (FBGT) and Filtering After GAN Training (FAGT). The CosSIF algorithm is employed to determine the similarity between two sets of images. For instance, in a dataset consisting of two classes, A and B, CosSIF calculates the similarity of each image from class A with all the images in class B. The resulting similarity scores generated by CosSIF are then used by FBGT or FAGT to filter out the most similar or dissimilar images. FBGT involves removing real images from the minority class that exhibit visual resemblance to images from other classes before incorporating them into the training dataset of a GAN. This ensures that the GAN does not learn certain features that contribute to generating visually similar images. However, implementing FBGT requires retraining the GAN with the filtered images. In contrast, FAGT operates on a pre-trained GAN, where similarity calculations are conducted between the synthetic images generated by the GAN and the real images used for training the GAN. The architecture of our proposed algorithm and filtering methods is illustrated in Fig. \ref{fig:pipeline}. To evaluate the effectiveness of our approaches, we perform experiments using modern transformers such as Vision Transformer (ViT) and Swin Transformer, as well as convolutional-based networks like ConvNeXt. The key contributions of our work in this paper can be summarized as follows:

\begin{itemize}
\item We propose CosSIF, an image similarity calculation algorithm with cosine similarity as its backbone, capable of identifying visually similar images of a specific class to images of another class/classes in a dataset.
\item We propose two filtering methods, FBGT and FAGT, to regulate GANs synthetic image generation capabilities in an effort to reduce low inter-class variability in medical image datasets.
\item We propose a reproducible train-test split for the HAM10000 dataset, which can facilitate the comparison of our proposed methods with future experiments conducted by others.
\item We experimentally demonstrate that the utilization of FAGT on the ISIC-2016 dataset surpasses the baseline method, MelaNet \cite{zunair2020melanoma}, in terms of sensitivity by 1.59\% and AUC by 1.88\%. Furthermore, the utilization of FBGT exceeds the baseline method, IRv2+SA \cite{datta2021soft}, in terms of recall by 13.75\%, and with the sole implementation of FAGT, achieves a maximum accuracy of 94.44\%.
\end{itemize}

The remaining sections of this paper are organized as follows: In Section \ref{sec:related_works}, we discuss related studies on class imbalance and low inter-class variation in medical image classification. Moreover, we explore the usage of cosine similarity in computer vision. In Section \ref{sec:proposed_method}, we present a comprehensive overview of our proposed CosSIF algorithm, as well as the FBGT and FAGT filtering methods. Furthermore, this section provides detailed descriptions of our selected GAN architecture and gives a brief overview of our chosen transformer and convolutional-based network models. In Section \ref{sec:experiments}, we give an overview of the utilized datasets, present the selected configurations for classifier and GAN training. Subsequently, we perform experiments by employing our proposed algorithm and filtering methods. In Section \ref{sec:results}, we conduct an ablation study of our experiments and compare the performance of our trained classifiers against baseline methods. Finally, Section \ref{sec:conclusion} presents the conclusions and a discussion on possibilities for future work.

\section{Related Work}
\label{sec:related_works}
Several studies have delved into a multitude of strategies to address class imbalance in medical image datasets. These approaches encompass oversampling techniques that involve either transformations or the implementation of generative adversarial networks (GANs) \cite{garcea2022data}. For instance, Zunair and Hamza \cite{zunair2020melanoma} employed CycleGAN, a GAN model consisting of dual-generator and discriminator modules, to effectively increase the representation of the minority class in the dataset \cite{zhu2017unpaired}. On the other hand, researchers such as Datta et al. \cite{datta2021soft} and Lan et al. \cite{lan2022fixcaps} opted for alternative transformation methods, adjusting image rotation and focus to diversify the dataset without resorting to GANs. While these research papers used different methods to address the class imbalance issue, they did not propose any solutions to handle the low inter-class variation in the generated synthetic datasets. In this paper, we aim to address this issue by eliminating images that contain limited distinguishable features during the oversampling process.

Identifying visually similar images begins with mathematically calculating the similarity between two images, preferably in a higher dimension. Multiple formulas exist for this task, including Mean Square Error (MSE), Cosine Similarity, and Euclidean Distance. In the field of computer vision, the use of cosine similarity is fairly prevalent for calculating the similarity between images. Ilham et al. \cite{ilham2020image} used cosine similarity to compare the vector representation of a query image with the vector representations of all the images in the database. Similarly, Kaur et al. \cite{kaur2013image} proposed a content-based image retrieval system (CBIR)  to assist dermatologists in diagnosing skin diseases. The system utilized various techniques such as feature extraction, similarity matching, and cosine similarity to retrieve the most similar images to the query image. Tao et al. \cite{tao2017image} introduced Saliency-Guided Constrained Clustering approach with cosine similarity (SGC3) for image cosegmentation. This method employs cosine similarity to calculate the feature similarity between data points and its cluster centroid. Given the extensive utilization of cosine similarity in similarity calculations, as demonstrated by the mentioned authors, we also choose to employ it as the foundation of our similarity calculation algorithm. Similar to these authors, we transform the two input images into vectors and calculate their cosine similarity. However, as we need to calculate similarities for thousands of images, it becomes necessary to reduce the pixel dimensions of the input images. Therefore, we optimize our algorithm by reducing the number of pixels to 64x64. This adjustment significantly improves the computation time of our algorithm. Moreover, we utilize cosine similarity to compute the cosine distance, ensuring that the calculated distances are non-negative values.

% ========= Proposed Method ========= 
\section{Proposed Method}

\begin{figure*}[t!]
\centering
\includegraphics[width=0.92\textwidth]{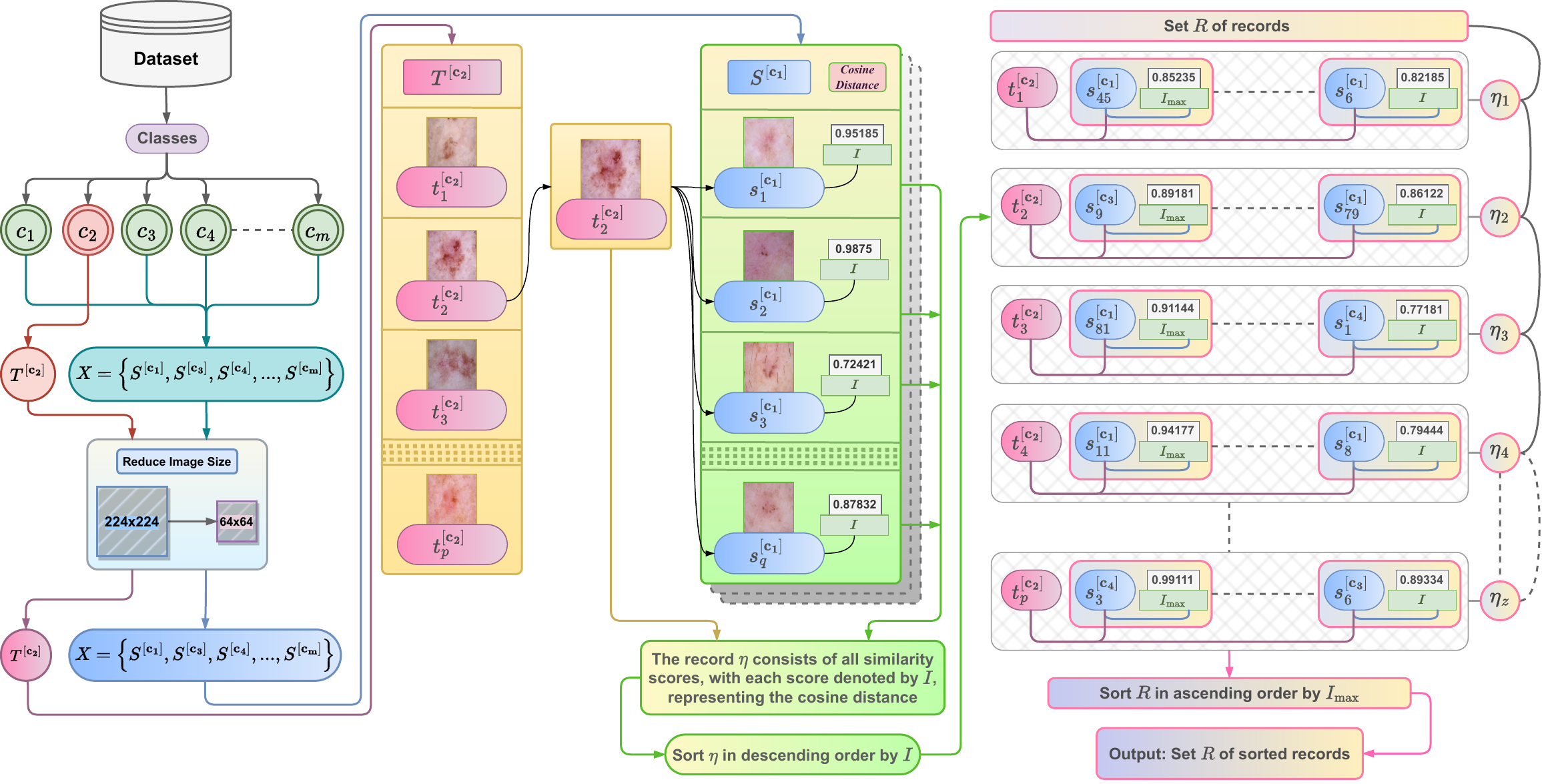}
\caption{The illustration portrays the step-by-step process of the CosSIF algorithm. It commences by selecting the target class ${T^{[c_2]}}$ and a set $X$ comprising secondary classes. Subsequently, the images within the selected classes undergo resizing to 64x64 pixels. Following this, similarity scores are calculated for each image in ${T^{[c_2]}}$ by comparing them to all images in $X$. For each image in ${T^{[c_2]}}$, a record $\eta$ is created to store individual similarity scores $I$ and their corresponding image identifiers. The record $\eta$ is then sorted in descending order, with the first entry representing the maximum similarity score $I_{max}$. Once the similarity calculation for all images in ${T^{[c_2]}}$ has been completed, the resulting set of records $R$ is obtained. Finally, $R$ is sorted in ascending order based on the maximum similarity score $I_{max}$, thereby concluding the similarity calculation process.}
\label{fig:CosSIF}
\end{figure*}

\label{sec:proposed_method}
This section describes the main components of our proposed algorithm and methods. It begins with a comprehensive overview of the CosSIF algorithm, followed by a detailed explanation of the FBGT and FAGT methods, along with a comparison between them. Next, the architecture of the GAN is described, along with a hybrid augmentation process designed specifically to suit the outlined GAN architecture. Finally, the section concisely discusses the transformers and convolutional-based network models used for training classifiers.

% ========= CosSIF =========
\subsection{CosSIF}
The details of the Cosine Similarity-based Image Filtering (CosSIF) algorithm are divided into several parts: class selection, image rescaling, similarity calculation, optimization, backbone, algorithm, filtering, and adaptability.

\subsubsection{Class Selection}
The CosSIF algorithm begins by selecting a class from a dataset, referred to as the target class. This target class acts as the anchor for similarity calculations, while the remaining classes are considered secondary classes. The target class is denoted as $\Tc$, and a secondary class is denoted as $\Sc$, where $c$ represents the class name. The total number of images in the target class is represented by $p$, and the total number of images in the secondary class is represented by $q$. Eq. \ref{eq:target_domain} and Eq. \ref{eq:secondary_domain} represent all the images within the $\Tc$ and $\Sc$ classes, respectively.

\begin{equation}
\label{eq:target_domain}
\Tc=\TcSet
\end{equation}

\begin{equation}
\label{eq:secondary_domain}
\Sc=\ScSet
\end{equation}

In the case of multiple secondary classes, they are represented as a set $X$. Eq. \ref{eq:secondary_domain_set} represents all the classes within the set $X$, where $m$ denotes the total number of secondary classes.

\begin{equation}
\label{eq:secondary_domain_set}
X=\SBoldD
\end{equation}

\subsubsection{Image Rescaling}
Upon selecting the target and secondary classes, all images belonging to these classes are resized to a smaller size, typically 64x64 pixels by default. This resizing step enables faster similarity calculations between images. To achieve even faster computation, the image size can be further reduced. Conversely, if a more detailed pixel-based computation is desired, the size can be increased beyond the default 64x64 pixels, albeit with an increase in computation time.

\begin{figure*}[h!]
    \centering
    \includegraphics[width=1.00\textwidth]{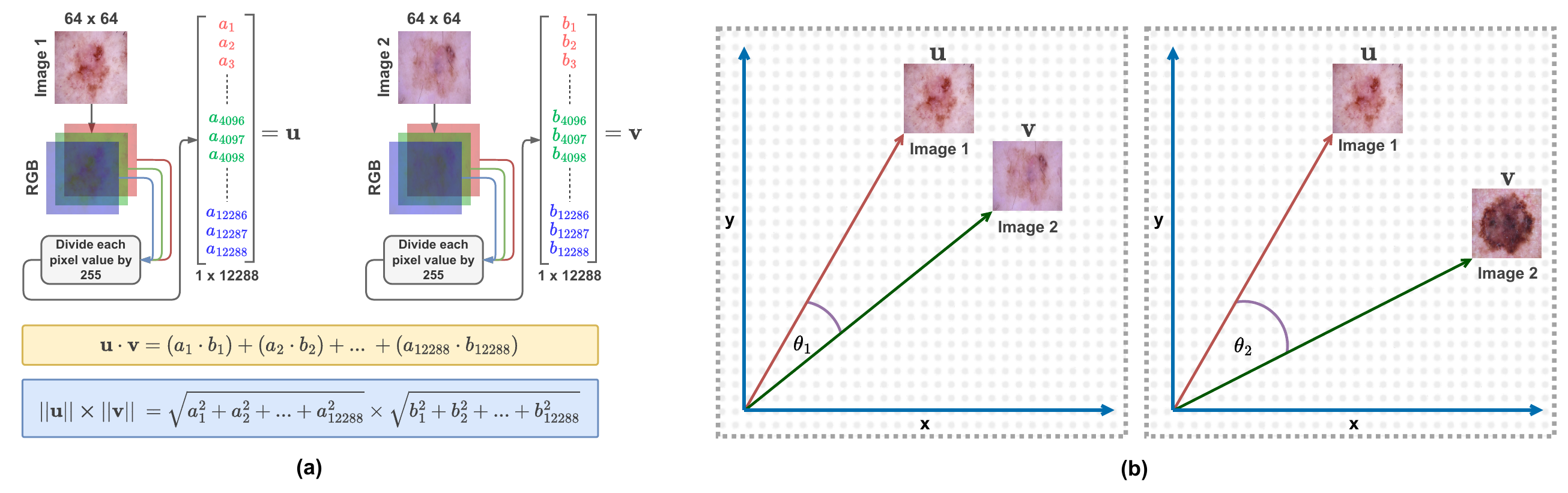}
    \caption{The illustration in (a) depicts the process of calculating the cosine similarity between two images. It begins by dividing a color image into its three RGB (red, green, blue) layers, where each layer is represented as a square matrix. Each layer contains different pixel values, which are then normalized by dividing each pixel by 255. Next, all layers are flattened into a vector. Considering that each image has a resolution of 64x64 pixels and consists of 3 layers, the resulting vector dimension becomes 1x12288. This procedure is repeated for both \textbf{Image 1} and \textbf{Image 2}, resulting in two vectors, $\mathbf{u}$ and $\mathbf{v}$, respectively. The cosine similarity between these two vectors is then calculated. The graphs in (b) illustrate that as the similarity between vectors $\mathbf{u}$ and $\mathbf{v}$ increases, the angle $\theta$ between them becomes smaller, and vice versa.}
    \label{fig:Cos_Sim_Cal}
\end{figure*}

\subsubsection{Similarity Calculation}
Following the image rescaling process, the algorithm proceeds to perform the similarity calculation. It starts by selecting an image from the target class $\Tc$ and calculates its similarity score, denoted as $I$, with all the other images in the secondary class $\Sc$. If there are multiple secondary classes, the similarity measure is computed for all images across the entire set $X$.

During the similarity calculation, a record denoted as $\eta$ is maintained for each image in $\Tc$, storing all the computed similarity scores along with the corresponding image identifiers (image and class names). The record $\eta$ is then sorted in descending order based on the individual similarity scores $I$. Therefore, for any image in $\Tc$, the first entry in the record $\eta$ contains the maximum similarity score, denoted as $I_{max}$, along with its associated image identifiers.

The algorithm iterates through all images in the target class $\Tc$ and records their similarities and corresponding image identifiers. Once the iteration process is complete, we obtain a set $R$ of records, as defined in Eq. \ref{eq:record_set}. It is important to note that the total number of records, denoted by $z$, in $R$, is equal to the total number of images, denoted by $p$, in $\Tc$.

\begin{equation}
\label{eq:record_set}
R = \{\eta_1, \eta_2, \ldots, \eta_z\}
\end{equation}

Finally, the set $R$ of records is sorted in ascending order based on the individual maximum similarity scores $I_{\text{max}}$. Fig. \ref{fig:CosSIF} provides a detailed illustration of the CosSIF algorithm.

\subsubsection{Optimization}
To tackle the issue of the record size growing excessively large as the number of images in $\Sc$ or $X$ increases, an optimization technique is introduced in the similarity calculation module. Rather than recording the similarity for each image in $\Tc$ with every other image in $\Sc$ or $X$, only a limited range of images with the highest similarity scores are recorded. However, the similarity calculation is still performed for all images in $\Sc$ or $X$. This approach effectively reduces the size of the record and addresses the scalability concern.

\subsubsection{Backbone}
The CosSIF algorithm analyzes images and computes their level of similarity. It utilizes cosine similarity to determine the degree of similarity between two images, as well as cosine distance to measure the positive distance between them.

Let's assume that $\mathbf{u}$ and $\mathbf{v}$ are two arbitrary vectors. The cosine similarity between the vectors is defined by Eq. \ref{eq:cos_sim}, where $\mathbf{u} \cdot \mathbf{v}$ represents the dot product of $\mathbf{u}$ and $\mathbf{v}$, and $\|\mathbf{u}\| \times \|\mathbf{v}\|$ denotes the product of their magnitudes. For a more detailed visual representation, refer to Fig. \ref{fig:Cos_Sim_Cal}.

\begin{equation}
\label{eq:cos_sim}
\begin{split}
    \text{cosine similarity $(\mathbf{u}, \mathbf{v})$}
    &=cos(\theta)
    =\frac{\mathbf{u} \cdot \mathbf{v}}{\mathbf{\|u\|} \times \mathbf{\|v\|}}\\
\end{split}
\end{equation}

The range of cosine similarity spans from -1 to 1. To convert this range to 0 to 1, the cosine distance is calculated. For cosine distance, a similarity value approaching 1 indicates more similar image pairs. Cosine distance is defined the Eq. \ref{eq:cos_dis}.

\begin{equation}
\label{eq:cos_dis}
    \text{cosine distance = 1 - cosine similarity}
\end{equation}

\subsubsection{Algorithm}
The detailed procedure of the CosSIF algorithm is presented in Algorithm \ref{alg:cal-similarity}. The input consists of the target class $\Tc$ and either a secondary class $\Sc$ or a set $X$ of secondary classes. The output is a set $R$ of records, sorted in ascending order, which holds the computed similarity scores for each image in $\Tc$ compared to all other images in $\Sc$ or $X$.

\begin{algorithm}[H]
\setstretch{1.00}
\caption{CosSIF}
\label{alg:cal-similarity}
\begin{algorithmic}[1]
\renewcommand{\algorithmicrequire}{\textbf{Input:}}
\renewcommand{\algorithmicensure}{\textbf{Output:}}
 
\REQUIRE Target class $\Tc=\TcSet$, secondary class $\Sc=\ScSet$ or a set $X$ of secondary classes.
 
\ENSURE Set $R$ of records, sorted in ascending order.
\STATE $R=\{\}$
\STATE Resize all images to 64x64.
\FOR{$\tci$ in $\Tc$}
    \STATE $\eta_1=\{\}$
    \IF {(secondary class != $X$)}  
        \STATE $X$ = $\{\Sc\}$
    \ELSE
        \FOR{$\Sc$ in $X$}
            \FOR{$\sci$ in $\Sc$}
                \STATE Calculate cosine similarity of $\tci$ and $\sci$. \ref{eq:cos_sim}
                \STATE Calculate cosine distance. \ref{eq:cos_dis}
                \STATE Similarity score $I$ = cosine distance
                \STATE Save record $\eta_1=\{\tci, \{\sci, I\}\}$
            \ENDFOR
        \ENDFOR   
        \STATE $\eta_1=\{\tci, \{\sci, I\}\}, \{s^{[c]}_{2}, I\}, \ldots\} $
        \STATE Sort $\eta_1$ in descending order by similarity scores.
        \STATE $\eta_1=\{\tci, \{s^{[c]}_{max}, I_{max}\}\}, \ldots\} $
    \ENDIF
    \STATE Append $\eta_1$ to set $R$.
\ENDFOR
\STATE Sort $R$ in ascending order by $I_{max}$.
\STATE Set $R=\{\eta_1, \eta_2, \ldots, \eta_z\}$
\end{algorithmic}
\end{algorithm}

\subsubsection{Filtering}
Once the similarity calculation is completed, the filtering process is initiated. To recap, the similarity calculation generates a set $R$ of records, as shown in Eq. \ref{eq:record_set}, which is subsequently sorted in ascending order. For each record $\eta$ in $R$, $I_{max}$ represents the maximum similarity score, and the associated image identifiers are used to identify the image in $\Tc$ that achieves this similarity score with an image in $\Sc$ or $X$. 

Since the set $R$ is sorted in ascending order by $I_{max}$, the first record, $\eta_1$, has the lowest $I_{max}$, while the last record, $\eta_z$, has the highest. Therefore, the process of filtering out the most similar images in begins with the $\eta_z$. From there, the filtering process gradually moves up the list of $\eta$ in $R$, with each subsequent image having a lower $I_{max}$ than the previous one. Conversely, to filter out the most dissimilar images, the filtering process starts from the $\eta_1$. In this case, the filtering process gradually moves down the list of $\eta$ in $R$, with each subsequent image having a higher similarity score than the previous one. 

In summary, the CosSIF algorithm enables the filtering of images based on their cosine similarity, identifying the most similar or dissimilar ones. However, the algorithm doesn't directly perform the filtering task; instead, it generates the essential information needed for filtering. The actual filtering is accomplished using our proposed FBGT and FAGT methods.

\subsubsection{Adaptability}
The CosSIF algorithm has been designed with future reusability in mind, allowing it to be applied to various tasks. To facilitate this adaptability, we have incorporated certain features that may not be immediately useful but could prove valuable in future works. One such feature relates to the similarity calculation process, where we provide the option to restrict the range of saved records. In the current implementation of our research, we have chosen to limit this range to only 1. This means that for each image in a selected target class, there is at most one image from the secondary class that is most similar. This suffices for our purposes, as the generated set $R$ of sorted records obtained from the CosSIF algorithm already gives us the most similar or dissimilar images that can be filtered from the target class. However, let's consider a scenario where it is necessary to know all possible similarities that each image in the target class shares with other images in the secondary class. In such cases, we can easily modify and reuse the CosSIF algorithm by adjusting the similarity range. By increasing the range, we can obtain the desired results and retrieve all the relevant similarities for each image. Therefore, the flexibility of the CosSIF algorithm allows it to be applied to a variety of tasks, and with slight modifications, it can accommodate different requirements in future applications.

\subsection{FBGT}
The Filtering Before GAN Training (FBGT) aims to eliminate real images from the minority class that display resemblances to images from other classes before employing them as the training dataset for a GAN. FBGT method commence by selecting the target and secondary classes. Here, the target class $\Tc$ represents the minority class within a given dataset, and the remaining classes are collectively referred to as a set $X$ of secondary classes. Then, the CosSIF algorithm is employed to calculate the similarity scores for each image in $\Tc$ with all other images in $X$. CosSIF generates a set $R$ of records which is shorted in ascending order. 

Following the completion of the similarity calculation and the generation of the set $R$ of records, the subsequent step in FBGT focuses on filtering images from the target class $\Tc$. In this step, the number of images to be filtered from $\Tc$ is determined by a hyperparameter denoted as $\alphaBold$, where $0 < \alpha < 1$. The value of $\alphaBold$ is calculated using the following formula:

\begin{equation}
    \label{eq:cal_alpha}
    \alpha = \frac{100 - \text{\% of images to be removed}}{100}
\end{equation}
The formula for calculating the number of filtered images, $f$, is given by:
\begin{equation}
    \label{eq:cal_f}
    f = \ceil*{p \times \alpha}
\end{equation}

where, the symbol $\ceil*{\ }$ represents the ceiling function, which rounds up the result of the multiplication to the nearest integer. The value of $f$ represents the threshold point, which indicates the number of images that have been filtered from $\Tc$. The newly filtered target class, $\Tf$, composed of $f$ images is given by:
\begin{equation}
    {\Tf=\{t^{[c]}_{1}, t^{[c]}_{2}, \ldots, t^{[c]}_{f}\}}
\end{equation}
$\Tf$ is the output of FBGT method. It contains the newly filtered images that are going to be used for oversampling. 

\subsection{FAGT}
The Filtering After GAN Training (FAGT) method calculates similarities between the synthetic images generated by a GAN and real images of the class on which the GAN was trained. In this method, the target class $\Tc$ consists of images that are synthetically generated via a trained GAN, while the secondary class $\Sc$ composed of real images, serves as the training dataset for that GAN. It is important to note that in FAGT, there is no possibility of having a set of secondary classes. Following the selection of $\Tc$ and $\Sc$, the FAGT method utilizes the CosSIF algorithm, leading to the generation of a set $\RBold$ of records. 

In the FAGT method, the process of filtering images from the target class $\Tc$ becomes more a bit more complex, compared to FBGT. Unlike FBGT, where the number of images in the filtered target class $f$ is not user-defined but rather calculated using Eq. \ref{eq:cal_alpha} and Eq. \ref{eq:cal_f}, in FAGT, the value of $f$ is determined by the user. This value represents both the number of images in the filtered target class $\Tf$ and the number of images required for oversampling. In FAGT, $f$ is considered a constant value. To control the output of the filtering process, the hyperparameter $\alpha$ is used to calculate the value of $p$, which denotes the number of synthetic images generated by a GAN. The formula for calculating $p$ is given by:

\begin{equation}
    \label{eq:cal_p}
    p = \ceil*{\frac{f}{\alpha}}
\end{equation} 

While $p$ is fixed in the FBGT method, it is variable in the FAGT method. This is because the quality of the synthetic images produced by a GAN can vary, leading to changes in the value of $p$. If a GAN produces synthetic images with fewer discriminative features compared to the real images, more images need to be filtered out from a larger set of images, resulting in an increased value of $p$. Conversely, if a GAN is capable of generating images with similar discriminative features compared to real images, then the value of $p$ decreases. This implies that fewer images need to be filtered out from a smaller set of synthetic images. Thus, in the FAGT method, $p$ behaves more like a hyperparameter. 

\begin{figure}[h]
    \centering
    \includegraphics[width=0.50\textwidth]{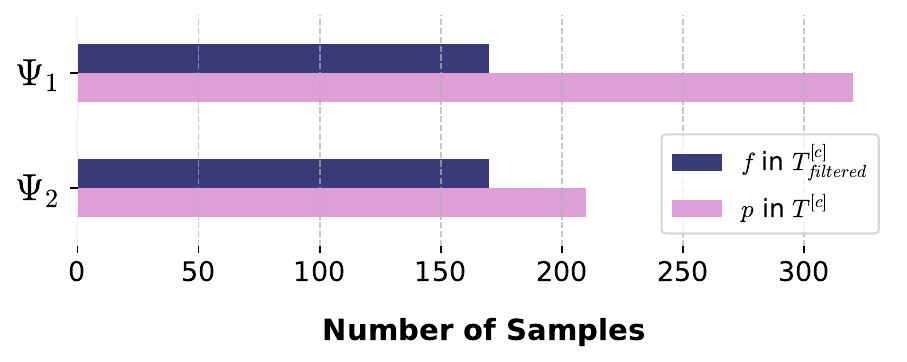}
    \caption{A visual representation of the correlation between the total number of images, $p$, in the target class $\Tc$, and the total number of images, $f$, in the filtered target class $\Tf$, employing the FAGT method, demonstrates that the extent of necessary filtering is influenced by the GAN's capacity to generate synthetic images that closely resemble the real images used to train the GAN.}
    \label{fig:FAGT_p-tuning}
\end{figure}

Fig. \ref{fig:FAGT_p-tuning} depicts this dependency in two setups, namely $\Psi_1$ and $\Psi_2$. In the first setup, $\Psi_1$, it is assumed that the GAN produces more random images with significant deviations from the real images, thereby necessitating a higher filtering requirement. Conversely, the second setup, $\Psi_2$, assumes that the GAN generates synthetic images that closely resemble the real images, resulting in a decreased need for filtering.

\subsection{Binary vs Multiclass Classification}
In both the FBGT and FAGT methods, it is possible to eliminate similar and dissimilar images from a calculated set $R$ of records. In binary classification, the removal of similar images from one class and dissimilar images from another class enhances the distinction between the two classes, resulting in improved filtering outcomes. However, in multiclass classification, it is crucial to avoid removing dissimilar images, as eliminating these images in relation to all other classes can lead to the loss of images that possess distinct features essential for accurate classification. Therefore, for multiclass classification, it is recommended to eliminate only the similar images.

\subsection{FBGT vs FAGT}
The FBGT and FAGT methods are two approaches that produces more robust oversampled datasets that reduce the issue of low inter-class variation. However, there are differences between these two methods that need to be considered. 

The FBGT method requires retraining the GAN with the newly filtered dataset, which can be a time-consuming process. Therefore, it may not be practical to use this method when dealing with multiclass classification problems that require oversampling for several classes. In contrast, the FAGT method can be applied to a pre-trained GAN, making it faster than the FBGT as it does not require retraining. However, the FAGT method requires filtering more images since $p$ is a variable for this method, resulting in longer computation time for filtering compared to the FBGT. 

In the FBGT method, both the target class and secondary class consist of real images. This implies that the set $\RBold$ of records obtained after the similarity calculation is universal and can be utilized later by anyone to filter out images. However, in the FAGT method, the target class is composed of synthetic images randomly generated by a trained GAN, which necessitates the recalculation of similarity each time the method is employed. As a result, the FBGT method is more efficient than the FAGT method when it comes to filtering images. 

Furthermore, it is essential to address a potential question regarding the FAGT method. Although similarities between real images of a specific class and real images from other classes are not directly calculated, the effectiveness of the method lies in the context of medical image datasets. Typically, all classes in such datasets consist of similar types of images (e.g., skin lesions, CT scans) but at different stages. When the GAN generates images that do not distinctly resemble the real images used during its training, these synthetic images may end up closely resembling images from other classes. As a result, the removal of synthetic images that deviate from the real images serves as a multipurpose filtering method. On one hand, it contributes to generating images with greater discriminative features, while on the other hand, it effectively addresses the issue of low inter-class variation.

\subsection{GAN Architecture}
The FBGT and FAGT methods are independent of GAN architecture, meaning that the filtering process remains constant regardless of any selected GAN framework. However, the choice of GAN architecture is often determined by the total image size in the dataset. Typically, training a GAN requires a large number of images. However, in medical image analysis, the minority class often consists of an extremely low volume of images, posing a challenge for the GAN to converge during training. Consequently, we tend to choose a GAN architecture that performs well with a small set of images.

In this paper, we use StyleGAN2-ADA as our GAN architecture. StyleGAN2-ADA builds upon the improvements of StyleGAN2, with the key enhancement being the addition of adaptive data augmentation (ADA) techniques. \cite{karras2020training}. This method allows for the generation of high-quality, diverse images even when the dataset is small or imbalanced, making it a valuable tool in medical image analysis and other applications where training data may be limited. StyleGAN2-ADA consists of two main parts: the generator and the discriminator. The generator's objective is to create realistic images, while the discriminator's goal is to differentiate between real and generated images. 

\subsubsection{Generator}
The generator $G$ consists of several key components: the mapping network, the synthesis network, and the Adaptive Instance Normalization (AdaIN) layers. 
The mapping network $h$ takes a latent code $\epsilon$ and maps it to a style vector $w$:
\begin{equation}
    w = h(\epsilon)
\end{equation}
The synthesis network $g$ takes the style vector $w$ and a noise tensor $n$, and generates an image $x$:
\begin{equation}
    x = g(w, n)
\end{equation}
Adaptive instance normalization (AdaIN) \cite{karras2020analyzing} is used to modulate the feature maps in the synthesis network with the style vector $w$. Given a feature map $F$ and the style vector $w$, AdaIN produces a styled feature map $F'$:
\begin{equation}
    F' = \text{AdaIN}(F, w)
\end{equation}
\subsubsection{Discriminator}
The discriminator $D$ is a convolutional neural network that classifies whether an input image $x$ is real or generated. It takes an image $x$ as input and outputs a scalar probability value $y$:

\begin{equation}
    y = D(x)
\end{equation}

\subsubsection{Adaptive Discriminator Augmentation}
In StyleGAN2-ADA, the discriminator is trained on both the real images and their augmented counterparts. The augmentation function $V$ takes an image $x$ and an augmentation parameter $\mu$ to produce an augmented image $x'$:

\begin{equation}
\label{eq:gan_ada}
    x' = V(x, \mu)
\end{equation}

\subsubsection{Loss}
Both the generator and discriminator losses are based on the binary cross-entropy loss function. The generator seeks to minimize its loss, which represents the difference between the discriminator's output on generated images and the target output. In essence, the generator aims to maximize the probability of the discriminator classifying the generated images as real:
\begin{align}
    L_G = -\mathbb{E}_{\epsilon \sim y(\epsilon)}[\log D(g(h(\epsilon), n))]
\end{align}
Here, $\epsilon$ is a random latent code sampled from the prior distribution $y(\epsilon)$, $h$ is the mapping network, $g$ is the synthesis network, $n$ is the noise tensor, and $D$ is the discriminator. $\mathbb{E}$ is expectation, which represents the average value of the expression inside the brackets. 

The discriminator aims to minimize its loss, which consists of two parts: the difference between the discriminator's output on real images and the target output, and the difference between the discriminator's output on generated images and the target output:
\begin{align}
    L_D &= -\mathbb{E}_{x \sim y_{\text{data}}(x)}[\log D(x)] \nonumber \\ &\quad - \mathbb{E}_{\epsilon \sim y(\epsilon)}[\log (1 - D(g(h(\epsilon), n)))]
\end{align}
Here, $x$ is an image sampled from the true data distribution $y_{\text{data}}(x)$, and the other variables have the same meaning as in the generator loss.

\subsection{Hybrid Augmentation}
Typically, oversampling via a GAN involves merging the real images with generated synthetic images after the GAN training. However, in some datasets, certain classes have an extremely low volume of training images. This makes the learning process of a GAN really difficult, even when using StyleGAN2-ADA.

The architecture of StyleGAN2-ADA includes a component known as adaptive discriminator augmentation (ADA), which is vital for training with a small number of images. During training, this component takes an input image $x$ and produces an augmented image $x'$, as expressed in Eq.\ref{eq:gan_ada}. To increase the variability of $x'$, it's important for the training dataset to contain sufficient variation. Therefore, we perform a minor oversampling of the minority classes by applying various transformations to the images before using them as a training dataset for StyleGAN2-ADA. These transformations include adjusting the focus, rotating the images, shifting their positions, and flipping them horizontally or vertically. The oversampling via transformations improves the variability of $x'$ during the process, which enhances the quality of the synthetically generated images produced by StyleGAN2-ADA. Refer to Fig. \ref{fig:pipeline} to visualize the oversampling via transformation of the minority class before utilizing them in the training dataset of a GAN.

\subsection{Model Architectures}
To assess the efficacy of our FBGT and FAGT methods, we employ pre-trained transformer and convolutional-based models. We train these models using the oversampled dataset, incorporating the FBGT and FAGT methods in some instances and excluding them in others for comparison purposes. For our experiments, we utilize pre-trained Swin Transformer \cite{liu2021swin}, Vision Transformer (ViT) \cite{dosovitskiy2020image}, and ConvNeXt \cite{liu2022convnet} models. We fine-tune these models by adapting their output layers to accommodate the classes within our dataset. 

The Swin Transformer, proposed by Liu et al. \cite{liu2021swin}, is a hierarchical transformer model specifically designed for computer vision tasks. It introduces a local representation to capture both local and global context, using a shifted window-based self-attention mechanism and a hierarchical architecture. The Vision Transformer (ViT), introduced by Dosovitskiy et al. \cite{dosovitskiy2020image}, applies the transformer architecture to computer vision tasks by dividing input images into patches and processing them as tokens with positional encodings. It has shown excellent performance on large-scale datasets but is known to be data-hungry. ConvNeXt, a model introduced by Liu et al. \cite{liu2022convnet}, demonstrates the potential of pure ConvNets by modernizing a standard ResNet to compete with the performance of Vision Transformers. 

\section{Experiments}
\label{sec:experiments}
In this section, extensive experiments are carried out to assess the performance of the proposed FBGT and FAGT methods, comparing them with strong baseline methods. 

\subsection{Datasets}
The effectiveness of FBGT and FAGT methods is analyzed using two datasets: ISIC-2016 and HAM10000. The ISIC-2016 dataset is employed to test both methods for binary classification, while the HAM10000 dataset is used for multiclass classification. Both datasets exhibit significant class imbalance with low inter-class variation, making them ideal choices for testing the filtering methods.  

\subsubsection{ISIC-2016 Dataset}
\begin{figure}[b!]
    \centering
    \includegraphics[width=0.48\textwidth]{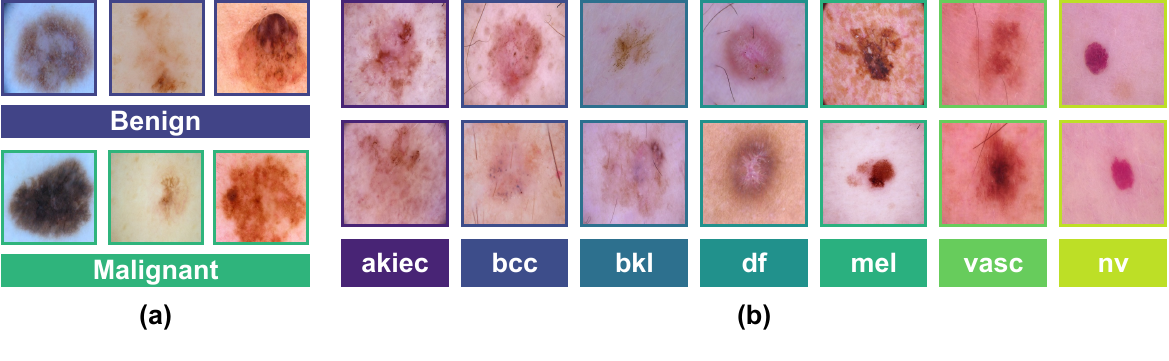}
    \caption{The visual depictions of images from various classes indicate low inter-class variation both in the ISIC-2016 dataset, as shown in (a), and the HAM10000 dataset, as shown in (b).}
    \label{fig:datasets_images}
\end{figure}

The ISIC-2016 Task 3 dataset contains 900 training and 379 testing dermoscopic images for skin lesion analysis and melanoma classification. With 173 malignant and 737 benign lesions in the training set, and 75 malignant and 304 benign lesions in the testing set, the dataset exhibits class imbalance \cite{gutman2016skin}. The images of malignant and benign lesions exhibit similar appearances, leading to low inter-class variation within the dataset, as observed in Fig. \ref{fig:datasets_images} (a). 
% Fig. \ref{fig:ISIC-2016_dataset} depicts the number of images in each class along with the train-test split.

\begin{figure}[b!]
    \centering
    \includegraphics[width=0.50\textwidth]{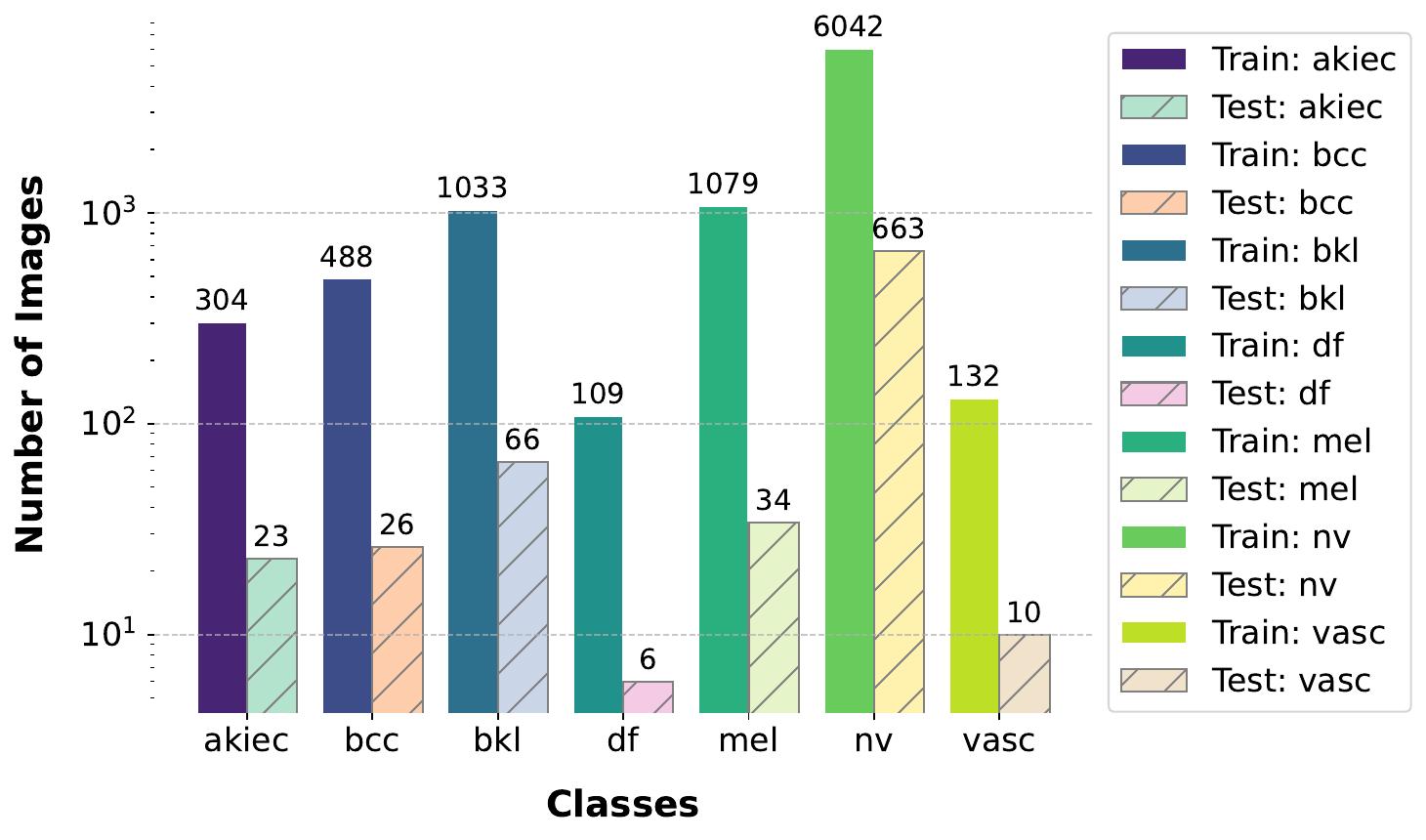}
    \caption{The bar plot provides a visual representation of the train-test split and the logarithmic scale depiction of the sample distribution across different classes within the HAM10000 dataset, revealing a significant class imbalance in both the training and testing datasets.}
    \label{fig:HAM10000_dataset}
\end{figure}

\subsubsection{HAM10000 Dataset}
The HAM10000 dataset consists of 10,015 clinical images of skin lesions, sourced from various locations worldwide and annotated by dermatologists. The dataset includes seven classes of skin lesions: actinic keratoses and intraepithelial carcinoma (akiec) with 327 images, basal cell carcinoma (bcc) with 514 images, benign keratosis-like lesions (bkl) with 1,099 images, dermatofibroma (df) with 115 images, melanoma (mel) with 1,113 images, melanocytic nevi (nv) with 6,705 images, and vascular lesions (vasc) with 142 images \cite{tschandl2018ham10000}. This distribution results in a highly unbalanced dataset. Moreover, the images also exhibit low inter-class variation, as can be seen in Fig. \ref{fig:datasets_images} (b).

The HAM10000 dataset does not include a predefined train-test split. This is particularly problematic as with no pre-defined split, it is difficult to compare the performance of our work with already existing state-of-the-art approaches. Some research papers focus on demonstrating high accuracy, which can sometimes involve manipulating the associated test data. Moreover, most work does not provide a reproducible train-test split, leading to difficulties in verifying the reported results.

Therefore, to address this issue, we employ a reproducible train-test split in our work. We partition the dataset into 9,187 images for training and 828 for testing. This partitioning process entails removing duplicates from the test set, which are composed of identical images with slight visual augmentations. Consequently, the training set contains these augmented images, while the test set is devoid of different augmentations of the same images. To perform the split, we use the scikit-learn train-test split library, providing a random state of 42 as an input parameter. This approach ensures consistency in the images within the training and testing sets for future experiments. Fig. \ref{fig:HAM10000_dataset} depicts the quantity of images in each class along with the distribution of the train-test split.

\begin{figure*}
    \centering
    \includegraphics[width=1.00\textwidth]{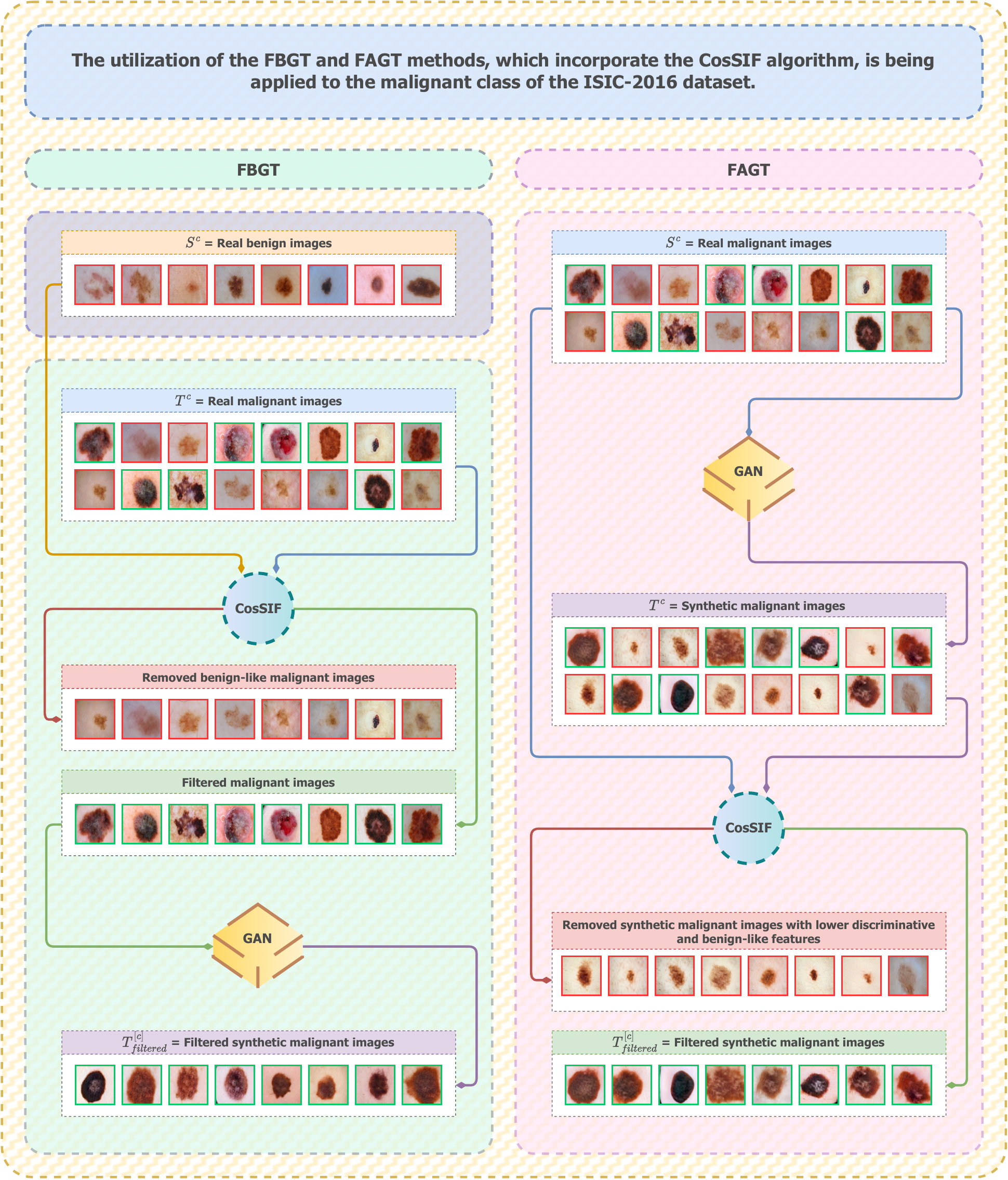}
    \caption{The illustration depicts the image filtering process of the malignant class in the ISIC-2016 dataset using our proposed FBGT and FAGT methods, which incorporate the CosSIF algorithm. The same filtering process is also applied to the benign class from the ISIC-2016 dataset, as well as the akiec, bcc, bkl, df, mel, and vasc classes from the HAM10000 dataset. In the visualization, the implication of FABT is shown for binary classification. In the case of multiclass classification, where there are multiple secondary classes, as observed in the HAM10000 dataset, CosSIF calculates the similarities against images of all secondary classes $X=\SBoldD$ with the chosen target class $\Tc$.}
    \label{fig:malignant_filtering}
\end{figure*}

\begin{table*}[ht]
    \centering
    \caption{The variation in the number of filtered images, denoted as $f$, from the total number of real/synthetic images, denoted as $p$, for three different values of $\alpha$ when implementing the FBGT or FAGT methods on the HAM10000 dataset.}
    \label{tab:HAM10000_alpha_tuning}

    \renewcommand{\arraystretch}{0.6}
    \resizebox{1.00\textwidth}{!}{
    \begin{tabular}{|c| c| c|c| c|c| c|c| c|c| c|c| c|c|}
        \toprule
        \multicolumn{1}{|c}{} & & \multicolumn{12}{c|}{Classes} \\
        \cmidrule{3-14}
        \multicolumn{1}{|c}{} & & 
        \multicolumn{2}{c|}{akiec} & 
        \multicolumn{2}{c|}{bcc} & 
        \multicolumn{2}{c|}{bkl} &
        \multicolumn{2}{c|}{df} &
        \multicolumn{2}{c|}{mel} &
        \multicolumn{2}{c|}{vasc}\\
        \cmidrule{3-14}
        \multicolumn{1}{|c}{} & $\alpha$ & Total $p$ & Filtered $f$ & Total $p$ & Filtered $f$ & Total $p$ & Filtered $f$ & Total $p$ & Filtered $f$ & Total $p$ & Filtered $f$ & Total $p$ & Filtered $f$ \\
        \midrule
        \midrule

        &
        $\alpha=$ 0.80 && 244 && 391 && 827 && 88 && 864 && 106 \\
        \cmidrule{2-2}
        \cmidrule{4-4}
        \cmidrule{6-6}
        \cmidrule{8-8}
        \cmidrule{10-10}
        \cmidrule{12-12}
        \cmidrule{14-14}

        &
        $\alpha=$ 0.85 & 304 & 259 & 488 & 415 & 1033 & 879 & 109 & 93 & 1079 & 918 & 132 & 113 \\
        \cmidrule{2-2}
        \cmidrule{4-4}
        \cmidrule{6-6}
        \cmidrule{8-8}
        \cmidrule{10-10}
        \cmidrule{12-12}
        \cmidrule{14-14}
        
        \multirow{-5}{*}{\rotatebox[origin=c]{90}{FBGT}}
        &
        $\alpha=$ 0.90 && 274 && 440 && 930 && 99 && 972 && 119 \\
        
        \midrule
        &
        $\alpha=$ 0.75 & 5624 && 5368 && 3241 && 7201 && 3142 && 7181 & \\
        \cmidrule{2-3}
        \cmidrule{5-5}
        \cmidrule{7-7}
        \cmidrule{9-9}
        \cmidrule{11-11}
        \cmidrule{13-13}
        
        &
        $\alpha=$ 0.80 & 5272 & 4218 & 5032 & 4026 & 3038 & 2431 & 6751 & 5401 & 2946 & 2357 & 6732 & 5386 \\
        \cmidrule{2-3}
        \cmidrule{5-5}
        \cmidrule{7-7}
        \cmidrule{9-9}
        \cmidrule{11-11}
        \cmidrule{13-13}
        
        \multirow{-5}{*}{\rotatebox[origin=c]{90}{FAGT}}
        &
        $\alpha=$ 0.85 & 4962 && 4736 && 2860 && 6354 && 2772 && 6336 & \\

        \bottomrule
    \end{tabular}
    }
\end{table*}

\subsection{Preprocessing}
In our experiment, we resize images from both datasets to a resolution of 256x256 pixels to ensure that the minority classes meet the criteria for being utilized as training datasets for the GAN. We opt to oversample both benign and malignant classes in the ISIC-2016 dataset and oversample all classes, except for melanocytic nevi (nv), in the HAM10000 dataset, as it already has a sufficient number of real images available and is the majority class.

\subsection{Dataset Filtering}
The dataset filtering process involves the utilization of the FBGT and FAGT methods on both the ISIC-2016 and HAM10000 datasets. Furthermore, as outlined in the methodology, we apply hybrid augmentation techniques to both datasets. This involves oversampling images through transformations, while simultaneously training GANs to generate synthetic images. We conduct a total of three primary experiments in our study. Experiment I utilizes the FBGT method, Experiment II employs the FAGT method, and Experiment III does not employ either the FBGT or FAGT methods.

\subsubsection{Experiment I}
In Experiment I, we apply the FBGT method to both benign and malignant (Fig. \ref{fig:malignant_filtering}, left) classes of the ISIC-2016 dataset and the akiec, bcc, bkl, df, mel, and vasc classes of the HAM10000 dataset. This involves conducting similarity calculations using CosSIF and filtering real images from the GAN training dataset that exhibit the highest similarity scores with images from other classes. We then perform minor oversampling via transformation with the newly filtered images associated with each class, followed by conducting GAN training for all the selected classes using the associated filtered GAN training datasets. Finally, we employ the trained GANs to generate synthetic images for each selected class, consequently resolving the issues of low inter-class variation. For a visual representation of this process, please refer to Fig. \ref{fig:pipeline}.

In the FBGT method, we have a hyperparameter called $\alpha$ that determines the number of images to be filtered from the real images. Instead of randomly selecting a number for filtering, we consider three specific values for $\alpha$. The variation in the number of filtered images for different $\alpha$ values when using the FBGT method can be observed in Table \ref{tab:HAM10000_alpha_tuning} and \ref{tab:ISIC-2016_alpha_tuning} for the HAM10000 and ISIC-2016 datasets, respectively.

\begin{table}[t!]
    \centering
    \caption{The variation in the number of filtered images, denoted as $f$, from the total number of real/synthetic images, denoted as $p$, for three different values of $\alpha$ when implementing the FBGT and FAGT methods on the benign and malignant classes of the ISIC-2016 dataset.}
    \label{tab:ISIC-2016_alpha_tuning}

    \renewcommand{\arraystretch}{0.1}
    \resizebox{0.49\textwidth}{!}{
    \begin{tabular}{|c|c|c|c|c|c|}
        \toprule
        \multicolumn{1}{|c}{} & & \multicolumn{4}{c|}{Classes} \\
        \cmidrule{3-6}
        \multicolumn{1}{|c}{} & & \multicolumn{2}{c|}{Benign} & \multicolumn{2}{c|}{Malignant} \\
        \cmidrule{3-6}
        \multicolumn{1}{|c}{} & $\alpha$ & Total $p$ & Filtered $f$ & Total $p$ & Filtered $f$ \\
        \midrule
        \midrule

        &
        $\alpha=$ 0.80 && 582 && 139 \\
        \cmidrule{2-2}
        \cmidrule{4-4}
        \cmidrule{6-6}

        &
        $\alpha=$ 0.85 & 727 & 618 & 173 & 148 \\
        \cmidrule{2-2}
        \cmidrule{4-4}
        \cmidrule{6-6}
        
        \multirow{-25}{*}{\rotatebox[origin=c]{90}{FBGT}}
        &
        $\alpha=$ 0.90 && 655 && 156 \\
        
        \midrule
        &
        $\alpha=$ 0.75 & 1441 && 1530 & \\
        \cmidrule{2-3}
        \cmidrule{5-5}

        &
        $\alpha=$ 0.80 & 1351 & 1081 & 1435 & 1148 \\
        \cmidrule{2-3}
        \cmidrule{5-5}
        
        \multirow{-25}{*}{\rotatebox[origin=c]{90}{FAGT}}
        &
        $\alpha=$ 0.85 & 1271 && 1350 & \\
        \bottomrule
    \end{tabular}
    }
\end{table}

\subsubsection{Experiment II}
In Experiment II, we apply the FAGT method to pre-trained GANs that are individually trained using the real images from the benign and malignant (Fig. \ref{fig:malignant_filtering}, right) classes of ISIC-2016, as well as the akiec, bcc, bkl, df, mel, and vasc classes of the HAM10000 dataset. By utilizing these pre-trained GANs, we generate synthetic images for each class. Subsequently, we perform similarity calculations to compare the generated synthetic images with the real training images. Based on the similarity results obtained from the calculations, we selectively remove synthetic images with lower discriminative features associated with the selected class. This process ensures that the filtered synthetic images closely resemble features of the real images used to train the GAN.

Like the FBGT method, the FAGT method also utilizes the hyperparameter known as $\alpha$ to determine the number of images to be filtered from the synthetic images. Similarly, we consider three specific values for $\alpha$, and the variations in the number of images can be observed in Table \ref{tab:HAM10000_alpha_tuning} and \ref{tab:ISIC-2016_alpha_tuning} for the HAM10000 and ISIC-2016 datasets, respectively. To visualize the internal processes of the FAGT method, please refer to Fig. \ref{fig:pipeline}.

\subsubsection{Experiment III}
In Experiment III, neither the FBGT nor FAGT methods are employed. Instead, we utilize the same pre-trained GAN as used in the FAGT method, but without implementing the similarity calculation and filtering process. This experiment is referred to as No-Filtering. Unlike the other experiments, No-Filtering does not involve any hyperparameter tuning. This experiment is solely conducted to analyze the efficacy of the FBGT and FAGT methods.

\subsection{Dataset Augmentation}
\begin{figure}[t!]
    \centering
    \includegraphics[width=0.46\textwidth]{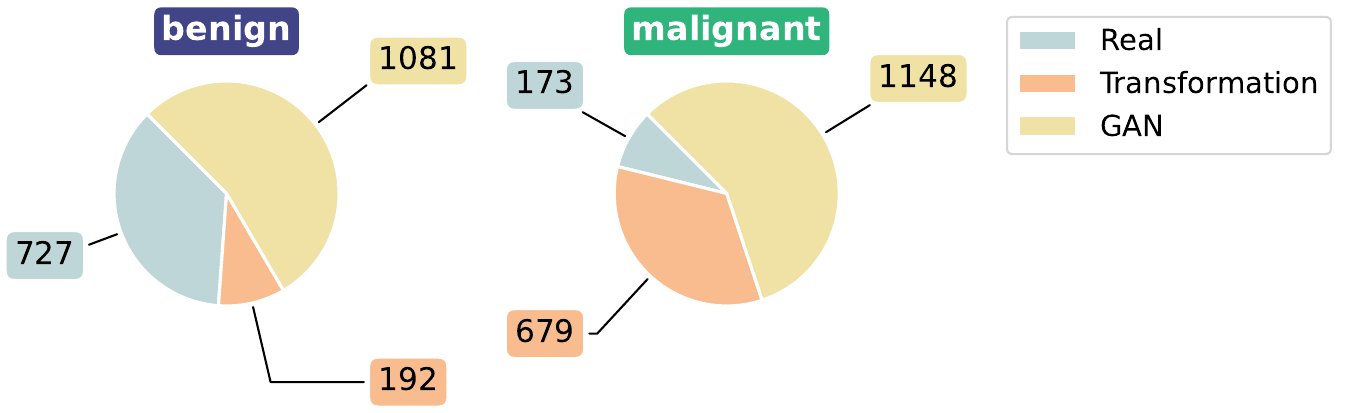}
    \caption{The pie charts show the composition of the augmented datasets, indicating the contributions of real, transformed, and synthetic images for the benign and malignant classes of the ISIC-2016 dataset.}
    \label{fig:ISIC-2016_augmented}
\end{figure}

\begin{figure}[t!]
    \centering
    \includegraphics[width=0.48\textwidth]{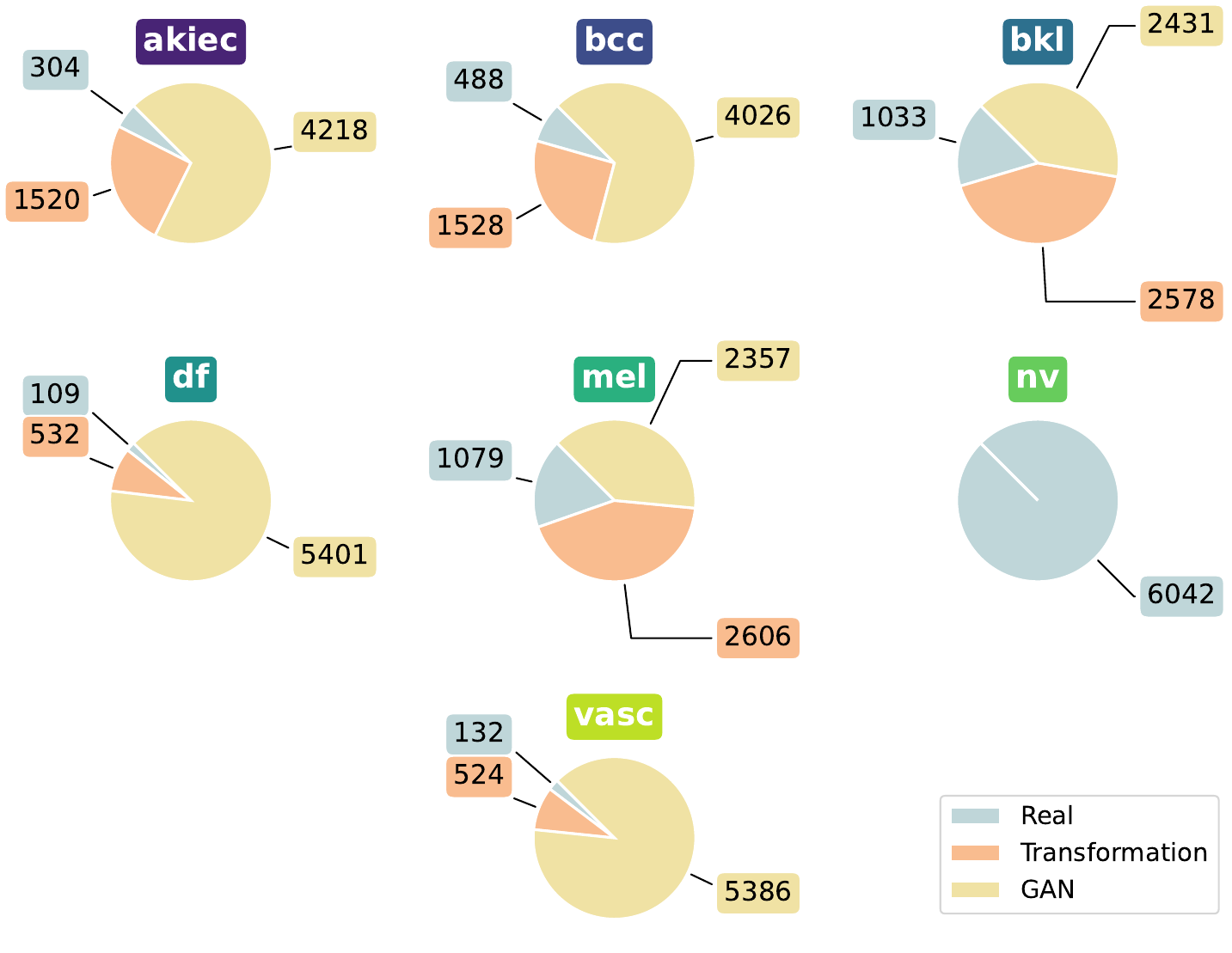}
    \caption{The pie charts visually represent the composition of the augmented dataset, showcasing the contributions of real, transformed, and synthetic images for all seven classes of the HAM10000 dataset.}
    \label{fig:HAM10000_augmented}
\end{figure}

Dataset augmentation is performed after the completion of dataset filtering. To achieve the final dataset augmentation for each class, we combine the real images with a batch of oversampled images obtained through transformations, as well as the synthetic images generated by GANs. During Experiment I, Experiment II, and Experiment III, the output is the final augmented dataset. The final augmented dataset, as shown in Fig. \ref{fig:ISIC-2016_augmented}, consists of 2000 images for both the benign and malignant classes, effectively addressing the class imbalance in the ISIC-2016 dataset. Similarly, in Fig. \ref{fig:HAM10000_augmented}, the final augmented HAM10000 dataset is displayed, with the akiec, bcc, bkl, df, mel, nv, and vasc classes each containing 6042 images. We chose this number as it matches the number of images in the overrepresented nv class, thus resolving the class imbalance present in the HAM10000 dataset by oversampling the remaining classes to this range.

Although the final augmented ISIC-2016 dataset contains 2000 images for each class and the HAM10000 dataset contains 6042 images for each class, it is important to note that these numbers are predetermined at the beginning of our experiments. There is a relationship in the FAGT method between the number of generated synthetic images and the number of filtered images. In this case, the number of filtered images is fixed and determined based on the size of the final augmented dataset. For example, in the ISIC-2016 dataset, we need to filter 1081 synthetic images from the benign class. This number is not randomly generated but derived from the size of the final augmented dataset. Therefore, if the final augmentation consists of 2000 images, with 727 real images and 192 images oversampled through transformations, the required number of synthetic images is calculated as (2000 - (727 + 192) = 1081). Thus, 1081 is the constant number used to control the output of the FAGT method. If the GAN generates more random images that don't resemble real images of the benign class, a larger pool of synthetic images must be generated, from which 1081 images are filtered. Conversely, if the GAN generates images that closely resemble real images of the benign class, a smaller pool of generated images is sufficient for filtering 1081 images.

\subsection{GAN Configuration}
As mentioned in the methodology section, we employ StyleGAN2-ADA as our GAN architecture. Each selected class for oversampling is trained using the same StyleGAN2-ADA configuration. In this configuration, we employ 400 kimg, which is equivalent to 400 epochs. Typically, StyleGAN2-ADA necessitates a larger number of epochs to generate realistic-looking synthetic images. However, for our experiment, we set kimg equal to 400 to accommodate our hardware constraints. Despite this limitation, we still acquire satisfactory synthetic images that fulfill our needs for analyzing the efficacy of our filtering methods.

\subsection{Training Classifiers}
As outlined in the methodology section, we utilize the Swin Transformer, ViT, and ConvNeXt models for training our classifiers. Specifically, we use the pre-trained versions of these models and fine-tune them using our final augmented datasets. To facilitate training, we resize the images in the final augmented datasets to 224x224 pixels. For optimization during training, we use AdamW, which is an algorithm designed for training deep learning models that extends the Adam optimizer to include weight decay regularization. We use a learning rate of $5e^{-5}$ during training for all these models. 

\subsection{Evaluation Metrics}
In classification tasks, it is essential to select appropriate evaluation metrics to accurately assess model performance. This section discusses various evaluation metrics, including recall, F1-score, sensitivity, accuracy, and AUC, which are employed to analyze the performance of modern transformer and convolutional-based network models and evaluate the efficacy of the FBGT and FAGT methods.

\subsubsection{Recall}
Recall, also known as sensitivity, measures the proportion of actual positive instances that are correctly predicted as positive. For binary classification, recall can be defined as:
\begin{equation}
Recall = \frac{TP}{TP + FN}
\end{equation}
where $TP$ represents the number of true positives and $FN$ represents the number of false negatives. For multiclass classification, macro-average recall is utilized. It is computed by calculating the recall for each class individually, treating each distinct class as a positive class and the remaining classes as negative classes. Then, the average of these recall values is taken. The formula for macro-average recall is given by:
\begin{equation}
Recall_{macro} = \frac{1}{k} \sum_{i=1}^{k} \frac{TP_i}{TP_i + FN_i}
\end{equation}
where $k$ represents the number of classes.

\subsubsection{F1-score}
The F1-score, which is the harmonic mean of precision and recall, provides a balance between the two metrics and is especially useful when dealing with imbalanced datasets or when both false positives and false negatives are of concern. In multiclass classification, we use the macro-average F1-score, which is calculated by computing the F1-score for each class individually and then taking the average of these values. The formula for macro-average F1-score is given by:
\begin{equation}
F1\text{-}score_{macro} = \frac{1}{k} \sum_{i=1}^{k} 2 \cdot \frac{Precision_i \cdot Recall_i}{Precision_i + Recall_i}
\end{equation}
where $Precision = \frac{TP}{TP + FP}$. By using the macro-average F1-score, we can obtain an overall performance measure of the multiclass classification model, while taking into account the performance of each individual class.

\subsubsection{Accuracy}
Accuracy measures the proportion of correctly predicted instances over the total number of instances. For binary classification, accuracy can be defined as:
\begin{equation}
Accuracy = \frac{TP + TN}{TP + TN + FP + FN}
\end{equation}
For multiclass classification, accuracy can be calculated as:
\begin{equation}
Accuracy = \frac{\sum_{i=1}^{k} TP_i}{\sum_{i=1}^{k} (TP_i + TN_i + FP_i + FN_i)}
\end{equation}

\subsubsection{AUC}
The AUC is the area under the receiver operating characteristic (ROC) curve, which is a plot of the true positive rate (TPR) against the false positive rate (FPR) at different classification thresholds. In the context of multiclass classification, the AUC is commonly computed using the one-vs-rest (OVR) strategy. In the OVR approach, we treat each class as the positive class and the remaining classes as the negative class, and we compute the AUC for each class separately. Then, we take the average of these AUC values to obtain the overall AUC score. 

\begin{table*}[t!]
    \centering
    \caption{Performance analysis of the fine-tuned Swin Transformer, ViT, and ConvNeXt models with the application of FBGT and FAGT methods for different $\alpha$ values, compared to No-Filtering, on the ISIC-2016 dataset. The table presents results in terms of sensitivity, false negative (FN) count, and AUC for the positive class. Boldface numbers indicate the best performance. The symbol ($\uparrow$) indicates higher is better, and the symbol ($\downarrow$) indicates lower is better.}
    \centering
    \renewcommand{\arraystretch}{1.2}
    \resizebox{1.0\textwidth}{!}{
    \begin{tabular}
    {@{}
    c@{\hspace{20pt}} 
    l@{\hspace{100pt}}
    c@{\hspace{100pt}} 
    c@{\hspace{100pt}}
    c@{}}
        
        \toprule
        & & & & Positive Class \\
        \cmidrule{5-5}
        Model & Filtering Method & Sensitivity (\%) ($\uparrow$) & FN ($\downarrow$) & AUC (\%) ($\uparrow$) \\
        \midrule
        \midrule

        % ========== Swin-Transformer ==========
        & 
        FBGT ($\alpha=$ \textbf{0.80}) & 
        \textbf{66.67} & 
        \textbf{25} & 
        81.36 \\

        & 
        FBGT ($\alpha=$ 0.85) &  
        57.33 & 
        32 & 
        83.13 \\

        & 
        FBGT ($\alpha=$ 0.90) &  
        62.67 & 
        28 & 
        82.81 \\
        \cmidrule{2-5}

        &
        FAGT ($\alpha=$ \textbf{0.75}) &
        \textbf{62.67} & 
        \textbf{28} & 
        83.40 \\
        
        & 
        FAGT ($\alpha=$ 0.80) &  
        58.67 & 
        31 & 
        83.45 \\

        & 
        FAGT ($\alpha=$ 0.85) &   
        58.67 & 
        31 & 
        83.17 \\
        \cmidrule{2-5}

        \multirow{-8}{*}{\rotatebox[origin=c]{90}{Swin-Transformer}}
        & 
        \text{No-Filtering} &  
        57.33 & 
        32 & 
        80.68 \\

        \midrule
        % ========== ViT ========== 
        & 
        \text{FBGT ($\alpha=$ 0.80)} &  
        60.00 & 
        30 & 
        83.91 \\
        
        &
        \text{FAGT ($\alpha=$ 0.75)} & 
        \textbf{72.00} & 
        \textbf{21} & 
        82.47 \\

        \multirow{-3.0}{*}{\rotatebox[origin=c]{90}{ViT}} 
        & 
        \text{No-Filtering} & 
        53.33 & 
        35 & 
        82.94 \\
        
        \midrule
        % ========== ConvNeXt ==========            
        & 
        \text{FBGT ($\alpha=$ 0.80)} &  
        \textbf{58.67} & 
        \textbf{31} & 
        79.38 \\
        
        &
        \text{FAGT ($\alpha=$ 0.75)} & 
        57.33 & 
        32 & 
        80.01 \\
        
        \multirow{-3.0}{*}{\rotatebox[origin=c]{90}{ConvNeXt}}
        & 
        \text{No-Filtering} &  
        56.00 & 
        33 & 
        75.46 \\

        \bottomrule
    \end{tabular}
    }
    \label{tab:isic2016-performance-analysis}
\end{table*}

\begin{figure*}[t!]
    \centering
    \begin{subfigure}[t]{1.00\textwidth}
        \includegraphics[width=\textwidth]{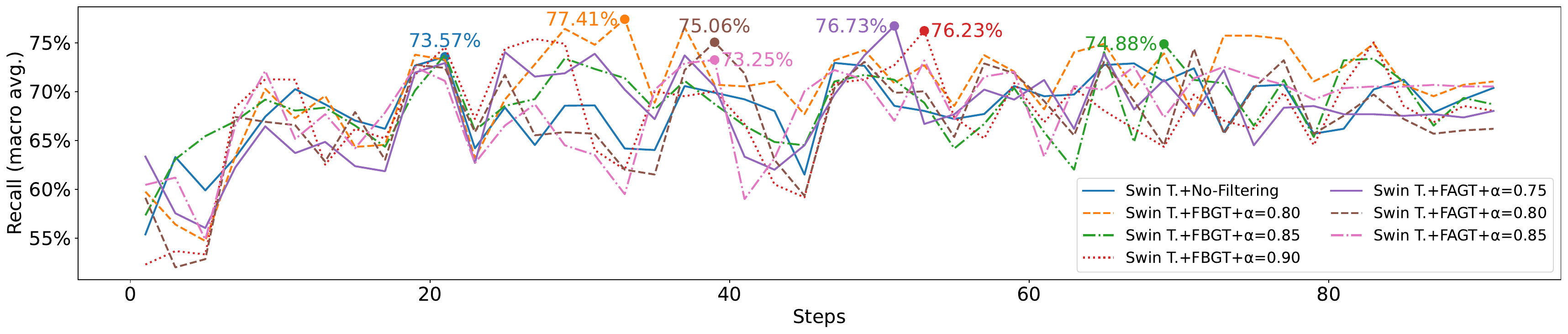}
    \end{subfigure}
    \begin{subfigure}[t]{1.00\textwidth}
        \includegraphics[width=\textwidth]{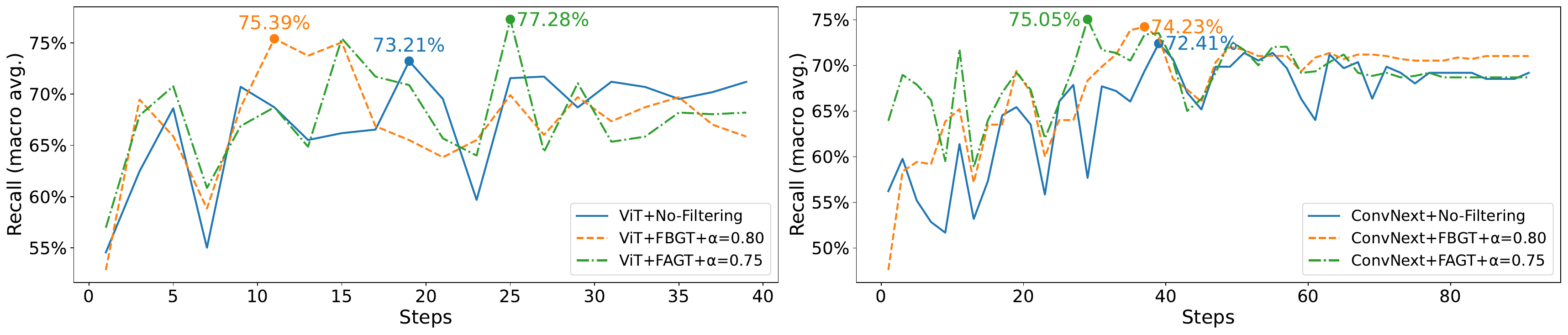}
    \end{subfigure}
    \caption{The graphs provide a comparison of the performance of fine-tuned Swin Transformer, ViT, and ConvNeXt models using FBGT and FAGT methods for different $\alpha$ values, in contrast to the No-Filtering approach, on the ISIC-2016 dataset. The maximum recall (macro avg.) serves as the pivotal criterion for selecting the optimal model during the classifier training process.}
    \label{fig:ISIC-2016-metric}
\end{figure*}

\subsection{Baselines}
Our baseline for the ISIC-2016 dataset is the results achieved by the MelaNet model, designed by Zunair and Hamza \cite{zunair2020melanoma}. However, due to an inconsistency in the reported sensitivity result in their paper, we recomputed this metric using the publicly available pre-trained MelaNet model. For the HAM10000 dataset, we use the results of the IRv2+SA model as our baseline, as reported by Datta et al. \cite{datta2021soft}. Both baselines demonstrate state-of-the-art classification performance. 

\subsection{Experimental Setup}
The experiments were conducted in two different setups with different hardware configurations. Setup 1, which employed a Linux server with 2-core Intel(R) Xeon(R) CPU @ 2.20GHz, 13 GB RAM, and 1x NVIDIA P100 16GB GPU was utilized for training StyleGAN2-ADA for different minority classes and for training classifiers for the HAM10000 dataset. Setup 2, which used a Windows machine with 6-core AMD Ryzen 5 5600H CPU @ 3.30GHz, 16 GB RAM, and 1x NVIDIA RTX 3060 6GB GPU, was utilized for implementing the FBGT and FAGT methods and for training classifiers for the ISIC-2016 dataset.

\section{Results}
\label{sec:results}
This section presents a thorough performance analysis of the FBGT and FAGT methods on each variation of the final augmented dataset. The analysis is conducted in two parts.  Firstly, we investigate the difference in performance while utilizing the FBGT and FAGT methods against No-Filtering. Secondly, we compare our best models with state-of-the-art baseline models. 

\subsection{Ablation Study} 
\begin{figure*}[t!]
    \centering
    \begin{subfigure}[t]{1.00\textwidth}
        \includegraphics[width=\textwidth]{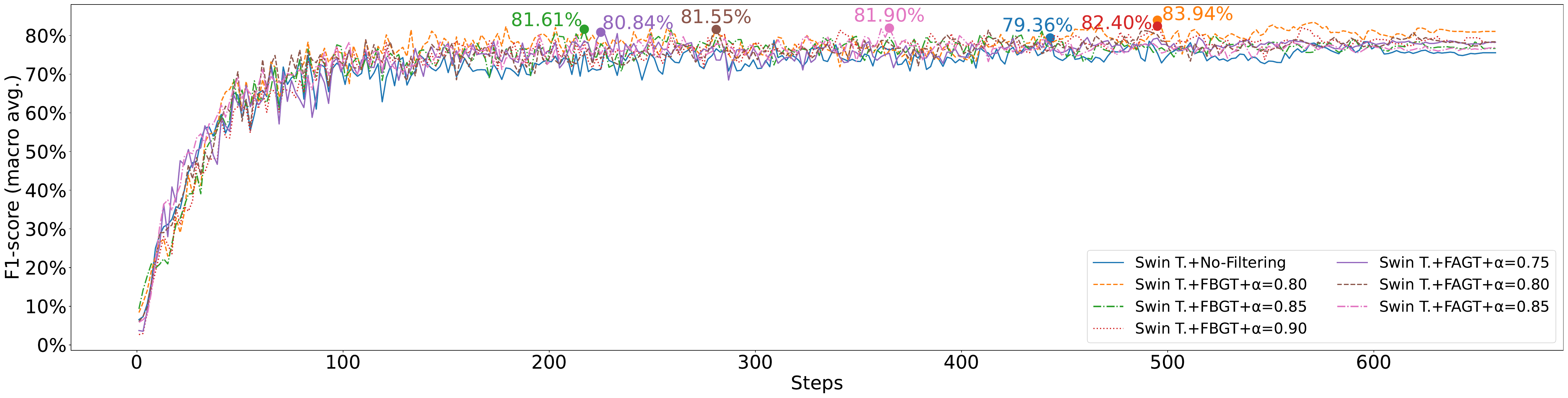}
    \end{subfigure}
    \begin{subfigure}[t]{1.00\textwidth}
        \includegraphics[width=\textwidth]{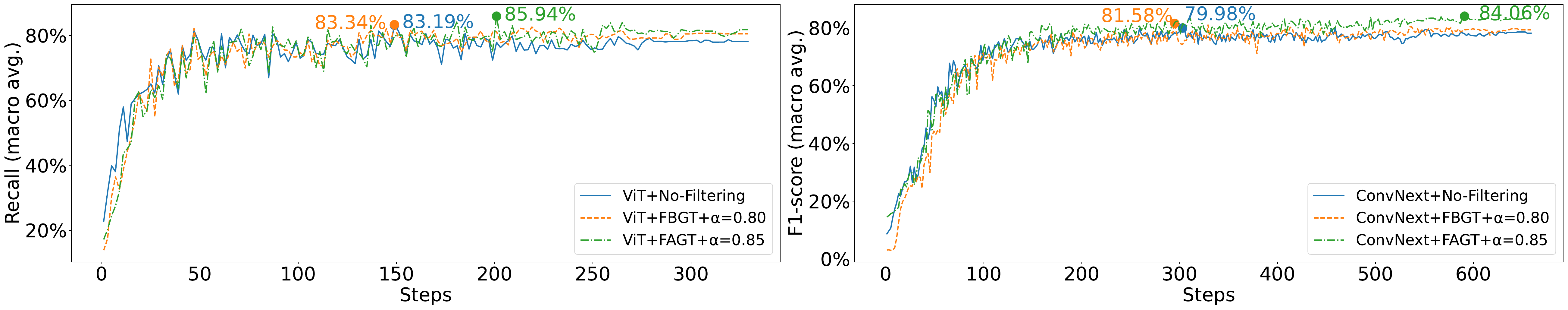}
    \end{subfigure}
    \caption{The graphs compare the performance of fine-tuned Swin Transformer, ViT, and ConvNeXt models using FBGT and FAGT methods for different $\alpha$ values, contrasting with the No-Filtering approach, on the HAM10000 dataset. The maximum recall (macro avg.) for the Swin Transformer and ConvNeXt, and the maximum F1-score for the ViT model, serve as pivotal criteria for selecting the optimal model during the classifier training process.}
    \label{fig:HAM10000-metric}
\end{figure*}

From the analysis presented in Table \ref{tab:isic2016-performance-analysis} and \ref{tab:ham10000-performance-analysis}, it is evident that our trained models exhibit improved performance across most evaluation metrics when utilizing either the FBGT or FAGT methods, as opposed to the No-Filtering approach, on both the ISIC-2016 and HAM10000 datasets. These performance gains are consistently observed across various distributions of augmented training datasets, each corresponding to a different $\alpha$ value. Furthermore, the graphs in Fig. \ref{fig:ISIC-2016-metric} and Fig. \ref{fig:HAM10000-metric} provide insights into the training phase of each model and the specific metric used to select the optimal variant during the classifier training process. 

\begin{table*}[t!]
    \centering
    \caption{Performance analysis of the fine-tuned Swin Transformer, ViT, and ConvNeXt models with the application of FBGT and FAGT methods for different $\alpha$ values, compared to No-Filtering, on the HAM10000 dataset. The table presents results in terms of recall, F1-score, accuracy, and average AUC. Boldface numbers indicate the best performance. The symbol ($\uparrow$) indicates higher is better.}
    \centering
    \renewcommand{\arraystretch}{1.2}
    \resizebox{1.0\textwidth}{!}{
    \begin{tabular}
    {@{}
    c@{\hspace{20pt}} 
    l@{\hspace{55pt}}
    c@{\hspace{55pt}}
    c
    c@{\hspace{55pt}}
    c@{\hspace{55pt}}
    c@{}}
        
        \toprule
        & & \multicolumn{2}{c}{Macro Average} & & & Average \\
        \cmidrule{3-4}
        \cmidrule{7-7}
        Model & Filtering Method & Recall (\%) ($\uparrow$) & F1-score (\%) ($\uparrow$) & & Accuracy (\%) ($\uparrow$) & AUC (\%) ($\uparrow$) \\
        \midrule
        \midrule
        
        % ========== Swin-Transformer ========== 
        & 
        FBGT ($\alpha=$ \textbf{0.80}) &  
        \textbf{81.82} & 
        \textbf{83.94} & &
        \textbf{94.08} &
        98.03 \\

        & 
        FBGT ($\alpha=$ 0.85) &  
        80.60 & 
        81.61 & &
        93.48 & 
        96.33 \\

        & 
        FBGT ($\alpha=$ 0.90) &  
        79.12 & 
        82.40 & &
        93.84 & 
        97.44 \\
        \cmidrule{2-7}

        & 
        FAGT ($\alpha=$ 0.75) & 
        81.90 & 
        80.84 & &
        93.36 &
        96.95 \\

        & 
        FAGT ($\alpha=$ 0.80) & 
        79.30 & 
        81.55 & &
        93.72 & 
        97.17 \\

        & 
        FAGT ($\alpha = 0.85$) & 
        \textbf{82.48} & 
        \textbf{81.90} & &
        \textbf{94.08} & 
        97.23 \\
        \cmidrule{2-7}

        \multirow{-8}{*}{\rotatebox[origin=c]{90}{Swin-Transformer}} 
        & 
        \text{No-Filtering} &  
        78.13 & 
        79.36 & &
        93.24 & 
        97.15 \\
        
        \midrule
        % ========== ViT ==========  
        & 
        \text{FBGT ($\alpha=$ 0.80)} &  
        83.34 & 
        81.85 & &
        92.39 & 
        97.67 \\

        &
        \text{FAGT ($\alpha=$ 0.85)} & 
        \textbf{85.94} & 
        \textbf{82.77} & &
        \textbf{93.00} & 
        98.27 \\

        \multirow{-3.0}{*}{\rotatebox[origin=c]{90}{ViT}} 
        & 
        \text{No-Filtering} & 
        83.19 & 
        81.12 & &
        91.79 & 
        96.95 \\
        
        \midrule
        % ========== ConvNeXt ==========
        & 
        \text{FBGT ($\alpha=$ 0.80)} &  
        81.38 & 
        81.58 & &
        93.60 & 
        97.65 \\
        
        &
        \text{FAGT ($\alpha=$ 0.85)} & 
        \textbf{81.80} & 
        \textbf{84.06} & &
        \textbf{94.44} & 
        97.54 \\

        \multirow{-3.0}{*}{\rotatebox[origin=c]{90}{ConvNeXt}} 
        & 
        \text{No-Filtering} &  
        79.86 & 
        79.98 & &
        92.63 & 
        97.32 \\
        
        \bottomrule
    \end{tabular}
    }
    \label{tab:ham10000-performance-analysis}
\end{table*}

\subsubsection{ISIC-2016 Dataset}
Table \ref{tab:isic2016-performance-analysis} and the graphs in Fig. \ref{fig:ISIC-2016-metric} offer an in-depth analysis of the performance of various models trained on the ISIC-2016 dataset. Regarding individual model performance, when employing the FBGT method, the Swin Transformer model with a hyperparameter value of $\alpha=0.80$ achieves the best performance, with a false negative (FN) count of 25, a sensitivity of 66.67\%, and a recall of 77.41\%. Similarly, when applying the FAGT method, the model with $\alpha=0.75$ demonstrates the best performance, resulting in an FN count of 28, a sensitivity of 62.67\%, and a recall of 76.73\%. These optimal $\alpha$ values ($\alpha=0.80$ for FBGT and $\alpha=0.75$ for FAGT) are also applied to the dataset distributions incorporating the filtering methods when training the ViT and ConvNeXt models. 

For the ViT model, applying the FAGT method with $\alpha=0.75$ leads to significant improvements, surpassing all other trained models with a sensitivity of 72.00\% and a FN count of 21. While the utilization of the FBGT method improves the ViT model's performance, yielding a sensitivity of 60\% and an FN count of 30, this improvement is comparatively less significant compared to FBGT. Similarly, the ConvNeXt model demonstrates performance enhancements with both the FBGT and FAGT methods, although the gains are not as substantial as those observed in the other models.

\subsubsection{HAM10000 Dataset}
Table \ref{tab:ham10000-performance-analysis} and the graphs in Fig. \ref{fig:HAM10000-metric} provide a performance analysis of models trained on the HAM10000 dataset. Concerning individual model performance, utilizing the FBGT method with $\alpha=0.80$ yields the best results for the Swin Transformer model, achieving a recall of 81.82\% and an F1-score of 83.94\%.  Similarly, when the FAGT method is employed with an $\alpha=0.85$, the best performance is observed, resulting in a recall of 82.48\% and an F1-score of 81.90\%. These optimal $\alpha$ values ($\alpha=0.80$ for FBGT and $\alpha=0.85$ for FAGT) are also applied to the dataset distributions incorporating the filtering methods when training the ViT and ConvNeXt models. 

In the case of the ViT model, utilizing the FAGT method with $\alpha = 0.85$  achieves the highest recorded recall of 85.94\% and the highest average AUC of 98.27\%. As for the ConvNeXt model, applying the FAGT method with $\alpha = 0.85$ results in an accuracy of 94.44\%, an F1-score of 84.06\%, and a recall of 81.80\%. This is by far the best-performing model when we consider F1-score and accuracy as our evaluation metrics. The use of the FBGT method does improve the performance of the ViT and ConvNeXt models as well. However, this improvement isn't as significant as using the FAGT method. 

\begin{table*}[t!]
    \centering
    \caption{Comparison of our best-performing model against the baseline, and various approaches by researchers on the ISIC-2016 dataset, focusing on key metrics such as AUC, sensitivity, and FN, where boldface numbers indicate the best performance. The symbol ($\uparrow$) indicates higher is better, and the symbol ($\downarrow$) indicates lower is better.}
    \renewcommand{\arraystretch}{1.1}
        \resizebox{1.0\textwidth}{!}{
        \begin{tabular}{
        @{}
        l@{\hspace{80pt}}
        c
        c@{\hspace{80pt}} 
        c@{\hspace{80pt}} 
        c@{}}
            
            \toprule
            & Positive Class \\
            \cmidrule{2-2}
            Method & AUC (\%) ($\uparrow$) & & Sensitivity (\%) ($\uparrow$) & FN ($\downarrow$) \\
            \midrule
            \midrule
    
            \text{Yu et al. \cite{yu2016automated} (\textit{without seg.})} &  
            78.20 & &
            42.70 & 
            -  \\
            
            \text{Yu et al. \cite{yu2016automated} (\textit{with seg.})} &  
            78.30 & &
            54.70 & 
            -  \\
    
            \text{Gutman et al. \cite{gutman2016skin}} &  
            80.40 & &
            50.70 & 
            -  \\
            
            \text{MelaNet \cite{zunair2020melanoma} (\textbf{Baseline})} &  
            81.18 & &
            70.67 & 
            22 \\
            \midrule
    
            \text{\textbf{ViT+FAGT ($\alpha=0.75$)}} &  
            \textbf{82.47} & &
            \textbf{72.00} & 
            \textbf{21}  \\
        
            \bottomrule
        \end{tabular}
        }
        \label{tab:isic2016-comparison}
\end{table*}

\begin{table*}[t!]
    \centering
    \caption{Comparison of our best-performing models against the baseline, and various approaches by researchers on the HAM10000 dataset, focusing on key metrics such as accuracy and recall, where boldface numbers indicate the best performance. The symbol ($\uparrow$) indicates higher is better.}
    \renewcommand{\arraystretch}{1.1}
        \resizebox{1.0\textwidth}{!}{
        \begin{tabular}{@{}
        l@{\hspace{145pt}}
        c@{\hspace{145pt}} 
        c@{}}
            
            \toprule
            & & Macro Average \\
            \cmidrule{3-3}
            Method & Accuracy (\%) ($\uparrow$) & Recall (\%) ($\uparrow$) \\
            \midrule
            \midrule
    
            \text{Khan et al. \cite{khan2019multi}} &  
            89.80 &
            $-$ \\
    
            \text{Onur Sevli \cite{sevli2021deep}} &  
            91.51 &
            $-$ \\
    
            \text{Chaturvedi et al. \cite{chaturvedi2020multi}} &  
            93.20 &
            $-$ \\
    
            \text{Afza et al. \cite{afza2022multiclass}} &  
            93.40 &
            $-$ \\

            \text{IRv2+SA \cite{datta2021soft} (\textbf{Baseline})} &  
            93.48 &
            71.93 \\
            \midrule
    
            \text{\textbf{Swin T.+FBGT ($\alpha=0.80$)}} &  
            \textbf{94.04} & 
            \textbf{81.82} \\
            \midrule
    
            \text{\textbf{ConvNeXt+FAGT ($\alpha=0.85$)}} &  
            \textbf{94.44} & 
            \textbf{81.80} \\
        
            \bottomrule
        \end{tabular}
        }
        \label{tab:ham10000-comparison}
\end{table*}

\subsection{Comparison Against Baselines}
In this section, we compare our best-performing models against strong baseline models. As mentioned before, we choose the baseline model MelaNet \cite{zunair2020melanoma} for both the ISIC-2016 dataset and the baseline IRv2+SA \cite{datta2021soft} HAM10000 datasets. Furthermore, we compare our models with other strong models proposed by various researchers. The overall comparisons for the ISIC-2016 and HAM10000 datasets can be visualized in Tables \ref{tab:isic2016-comparison} and \ref{tab:ham10000-comparison}.

\subsubsection{ISIC-2016 Dataset}
Our best-performing model trained on the ISIC-2016 dataset is the ViT model, utilizing the FAGT method with $\alpha=0.75$. This model surpasses the baseline method, MelaNet, by 1.59\% in sensitivity and 1.88\% in AUC. Our baseline remained unmatched for pure classification without any segmentation until the introduction of our proposed method, which outperforms it in every evaluation category. In Table \ref{tab:isic2016-comparison}, a detailed comparison is presented, showcasing the performance of our best-performing model against the baseline and other approaches proposed in different studies on the ISIC-2016 dataset.

\subsubsection{HAM10000 Dataset}
The ConvNeXt model, utilizing the FAGT method with $\alpha=0.85$, is our best-performing model trained on the HAM10000 dataset. It outperforms the baseline method, IRv2+SA, by 13.72\% in recall and 1.03\% in accuracy. Similarly, the Swin Transformer model, employing the FABT method with an $\alpha=0.80$, surpasses the baseline IRv2+SA by 13.75\% in recall and 0.60\% in accuracy. Furthermore, our model utilizes 42,294 training images, whereas the baseline IRv2+SA uses 51,699 images, making our approach more sample-efficient. Table \ref{tab:ham10000-comparison} provides a performance comparison of our best-performing models, against the baseline and other approaches proposed in different studies on the HAM10000 dataset.

\begin{figure*}[t!]
    \centering
    \begin{subfigure}[t]{0.226\textwidth}
        \includegraphics[width=\textwidth]{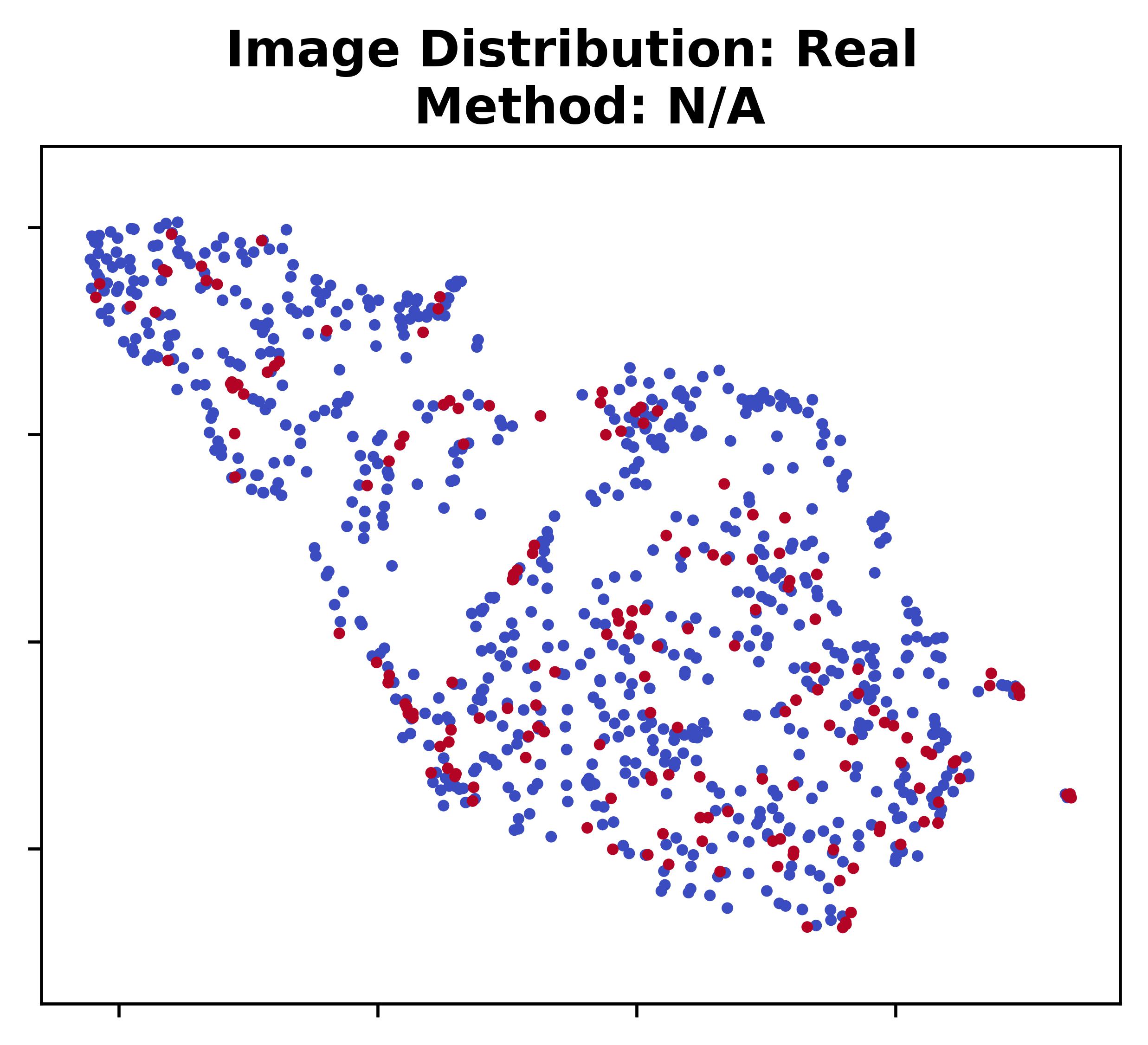}
        \caption{}
        \label{fig:11-a}
    \end{subfigure}
    \begin{subfigure}[t]{0.24\textwidth}
        \includegraphics[width=\textwidth]{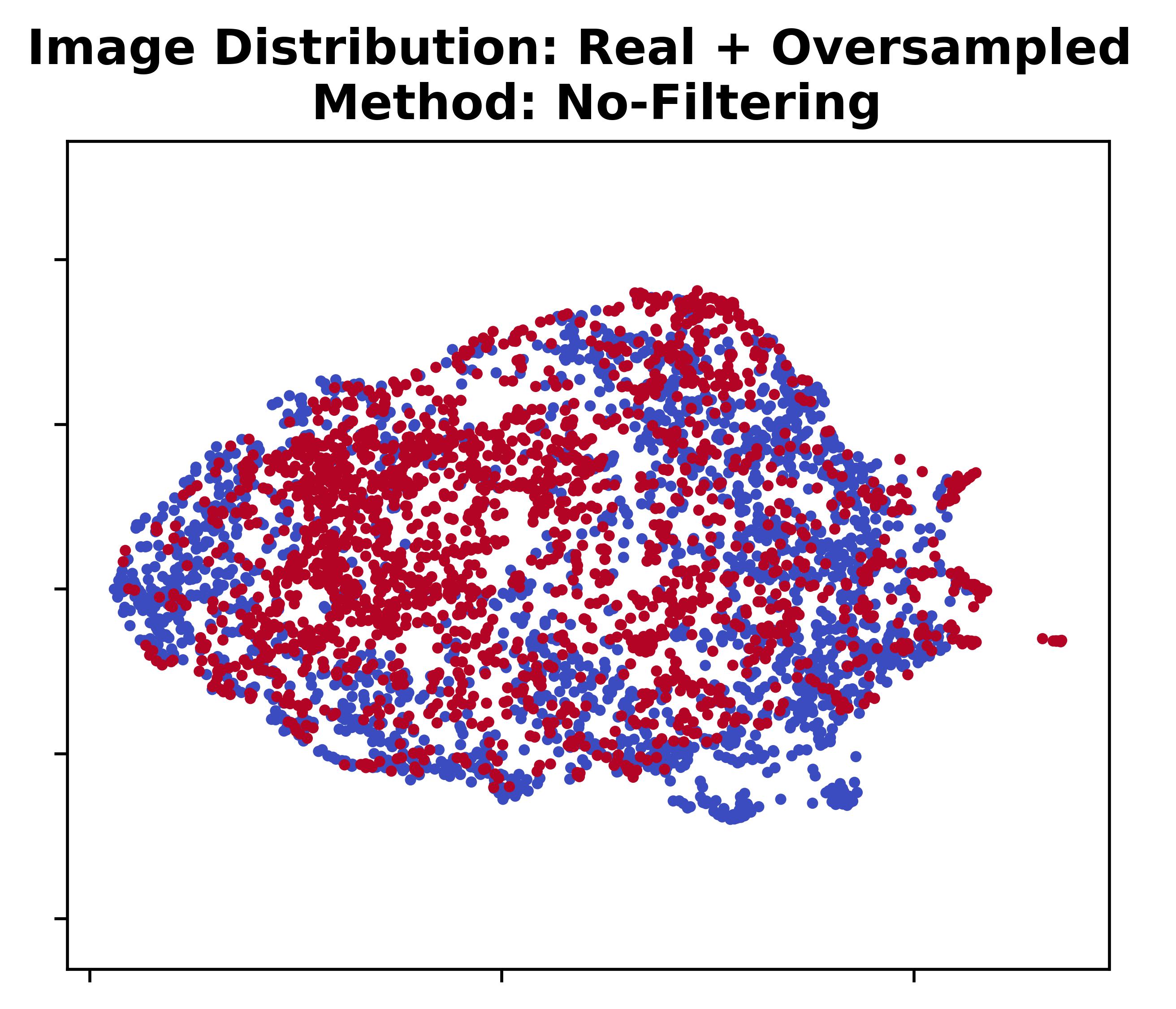}
        \caption{}
         \label{fig:11-b}
    \end{subfigure}
    \begin{subfigure}[t]{0.24\textwidth}
        \includegraphics[width=\textwidth]{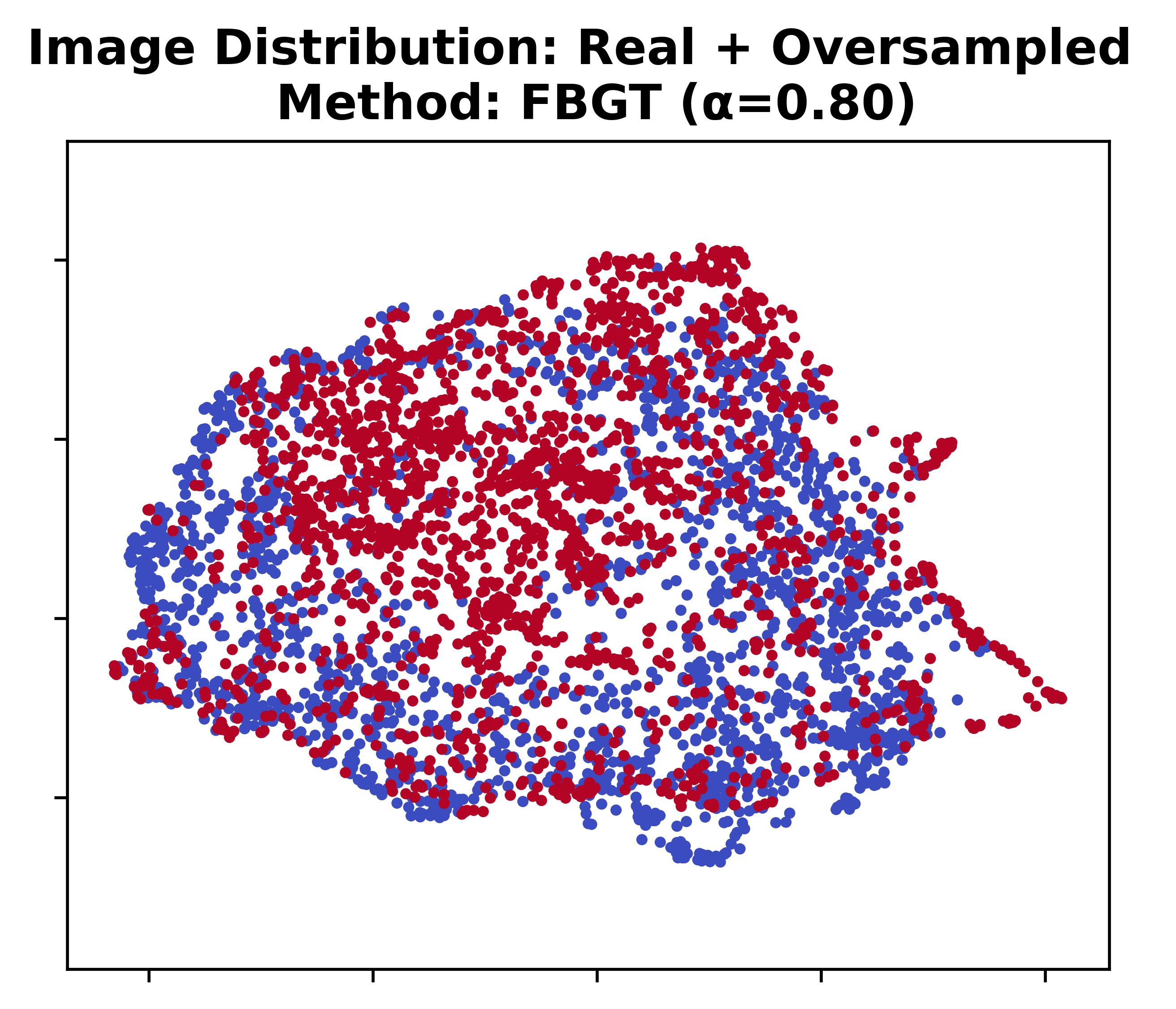}
        \caption{}
         \label{fig:11-c}
    \end{subfigure}
    \begin{subfigure}[t]{0.26\textwidth}
        \includegraphics[width=\textwidth]{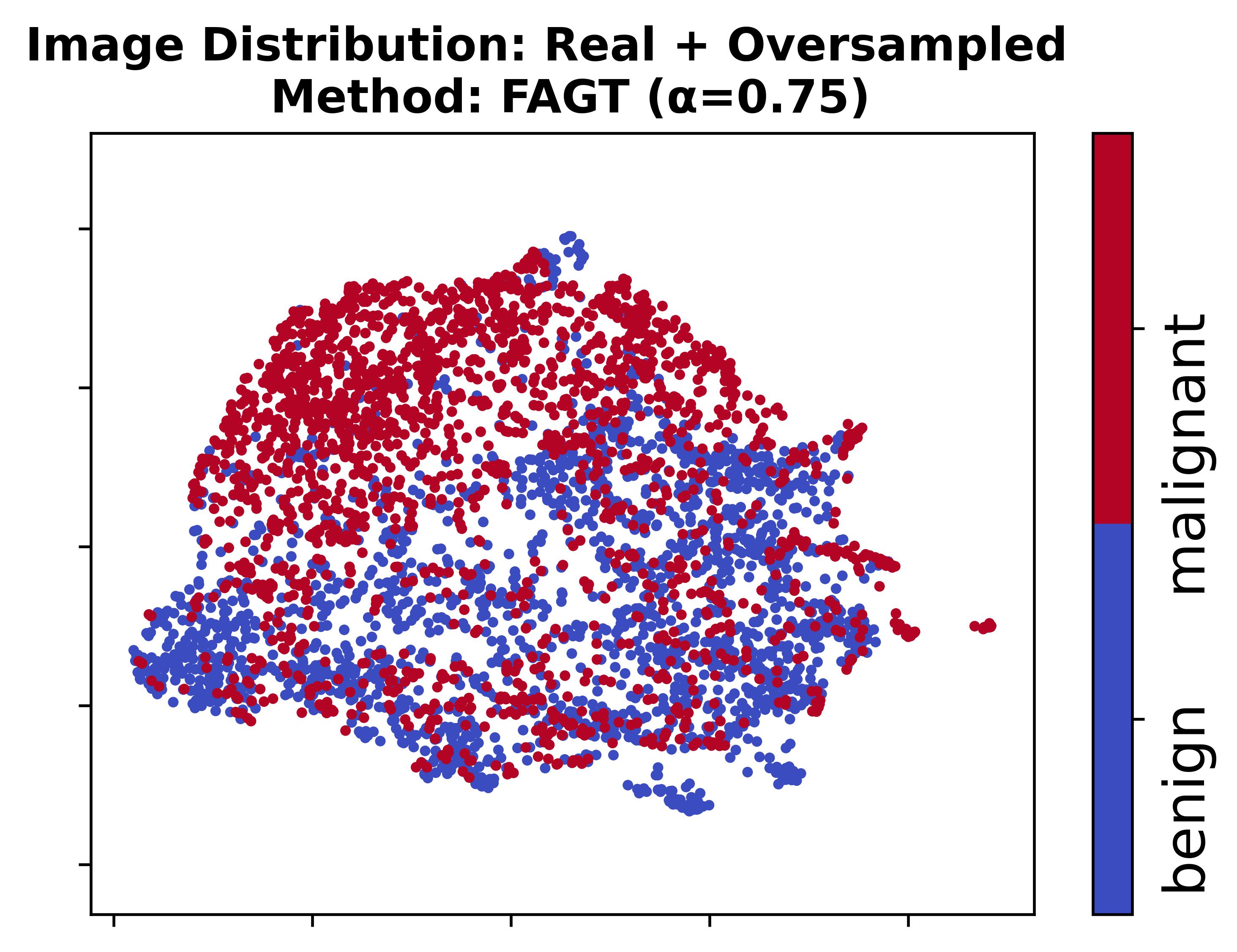}
        \caption{}
         \label{fig:11-d}
    \end{subfigure}
    \caption{The subfigures showcase the four variations of the ISIC-2016 dataset using 2D UMAP visualizations, offering a comparative view of the dataset's distribution and clustering patterns for each variation.}
    \label{fig:umap_ISIC-2016}
\end{figure*}

\begin{figure*}[t!]
    \centering
    \begin{subfigure}[t]{0.24\textwidth}
        \centering
        \includegraphics[width=\textwidth]{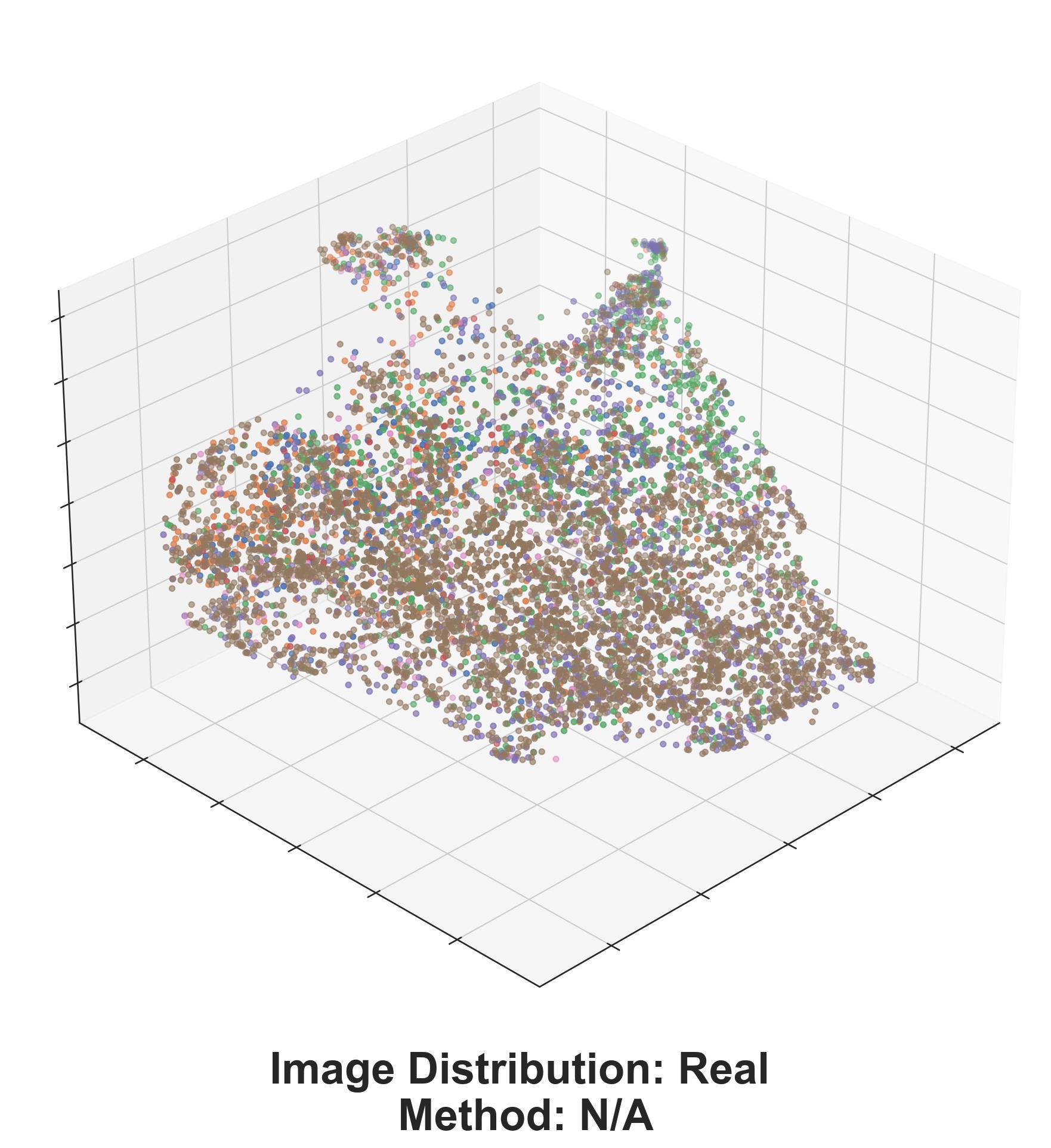}
        \caption{}
        \label{fig:12-a}
    \end{subfigure}
    \begin{subfigure}[t]{0.24\textwidth}
        \centering
        \includegraphics[width=\textwidth]{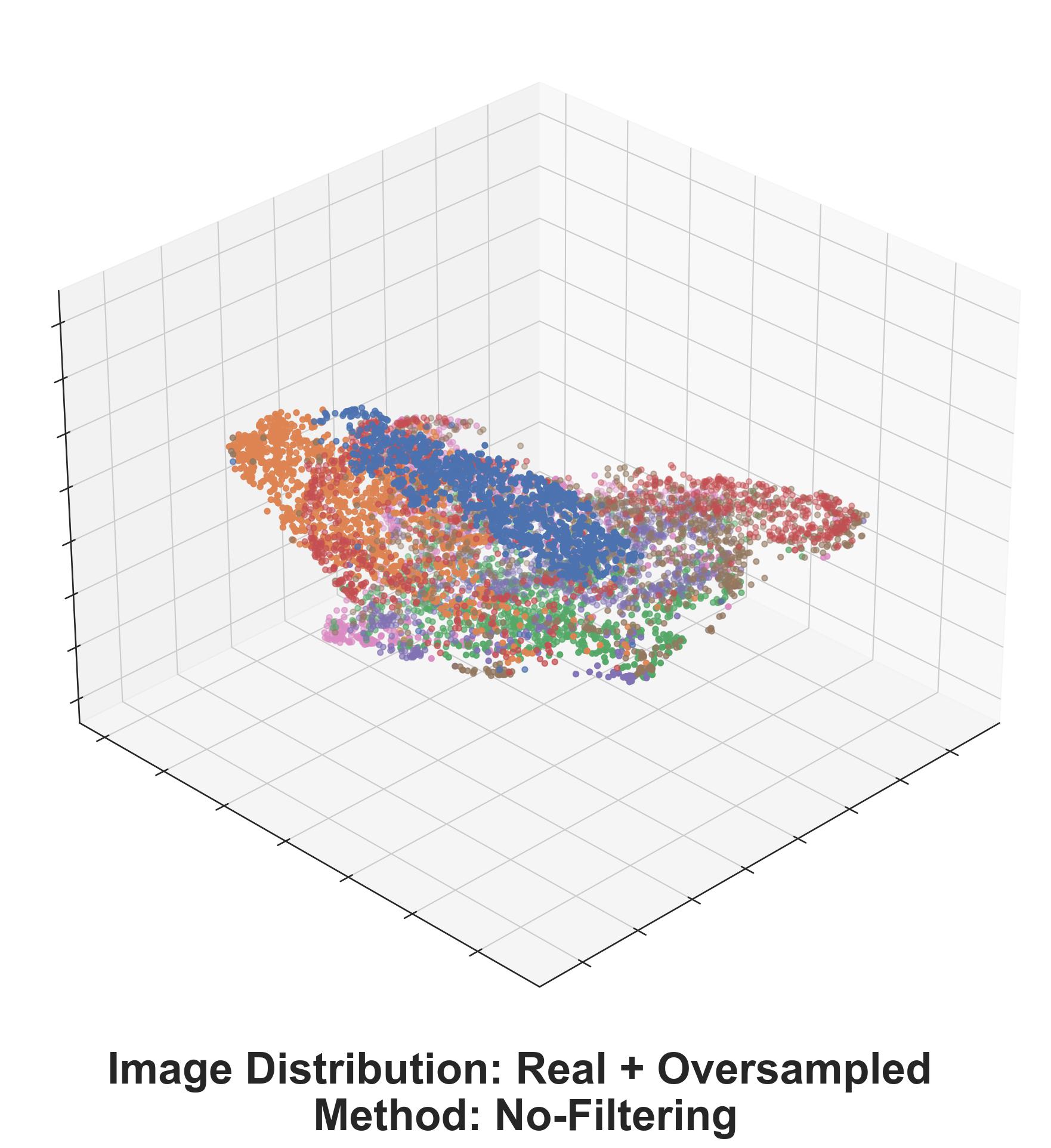}
        \caption{}
        \label{fig:12-b}
    \end{subfigure}
    \begin{subfigure}[t]{0.24\textwidth}
        \centering
        \includegraphics[width=\textwidth]{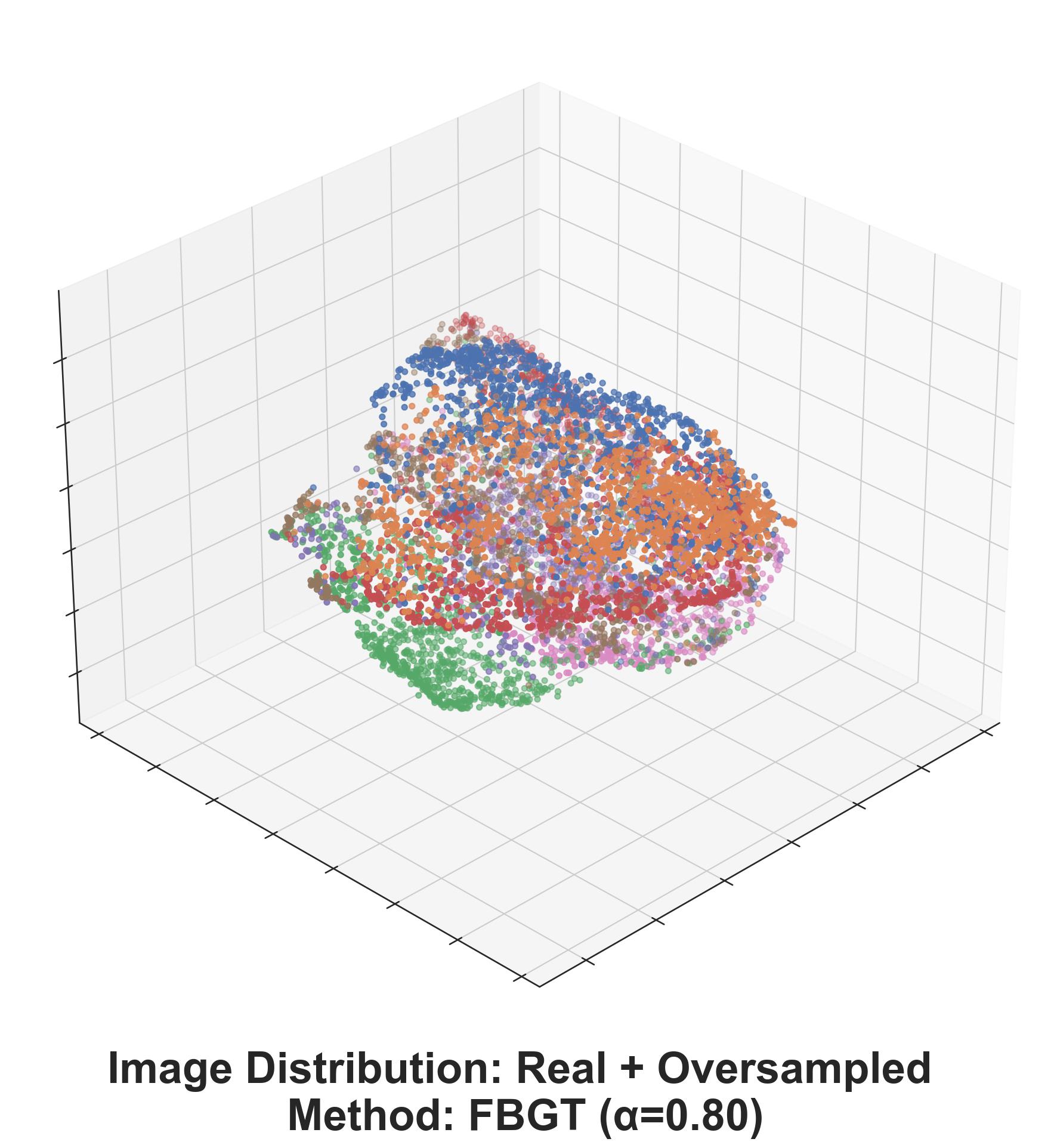}
        \caption{}
        \label{fig:12-c}
    \end{subfigure}
    \begin{subfigure}[t]{0.24\textwidth}
        \centering
        \includegraphics[width=\textwidth]{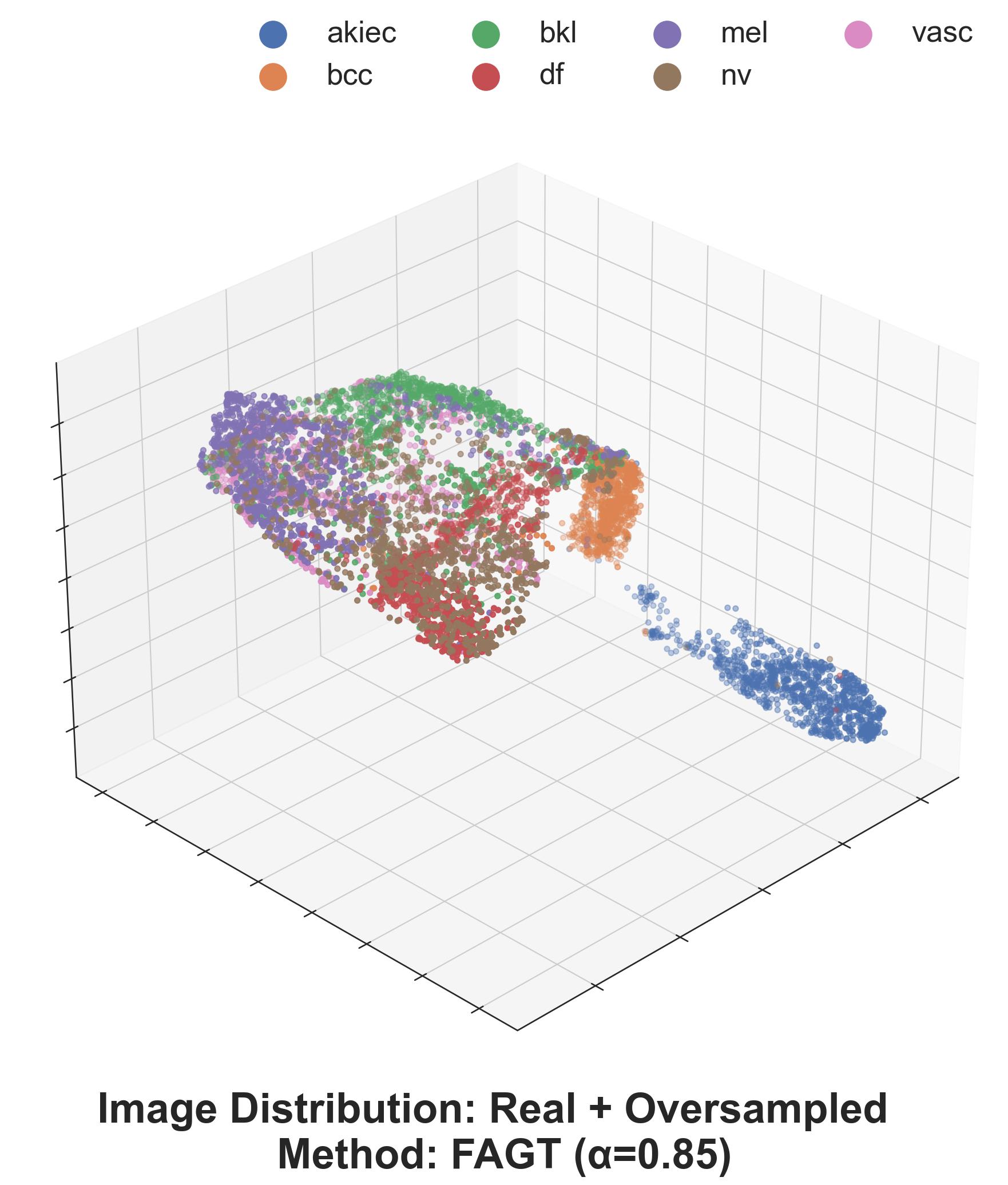}
        \caption{}
        \label{fig:12-d}
    \end{subfigure}
    \caption{The subfigures present 3D UMAP visualizations of the four different variations of the HAM10000 dataset, enabling a convenient comparison of the distribution and clustering patterns across each variation.}
    \label{fig:umap_HAM10000}
\end{figure*}

\subsection{Feature Visualization}
To visualize the data distribution, we utilize the Uniform Manifold Approximation and Projection for Dimension Reduction (UMAP) technique \cite{mcinnes2018umap}. We analyze four different variations of the ISIC-2016 dataset, which are presented in of Fig. \ref{fig:umap_ISIC-2016}, and four variations of the HAM10000 dataset, which are shown in Fig. \ref{fig:umap_HAM10000}.

Variation 1, depicted in Fig. \ref{fig:umap_ISIC-2016} (a) and Fig \ref{fig:umap_HAM10000} (a), represents the original dataset consisting of real images. Variation 2, displayed in Fig. \ref{fig:umap_ISIC-2016} (b) and Fig. \ref{fig:umap_HAM10000} (b), comprises unfiltered oversampled datasets that include real, transformed, and GAN-generated synthetic images. This variation corresponds to the No-Filtering experiment. In variations 3 and 4, illustrated in Fig. \ref{fig:umap_ISIC-2016} (c), Fig. \ref{fig:umap_ISIC-2016} (d), Fig. \ref{fig:umap_HAM10000} (c), and Fig. \ref{fig:umap_HAM10000} (d), we have filtered oversampled datasets that also incorporate real, transformed, and GAN-generated synthetic images. These variations utilize the FBGT and FAGT methods, and the composition of the datasets is determined by the hyperparameter $\alpha$. The selection of the dataset composition is based on the highest performance observed on the Swin Transformer classifier.

Fig. \ref{fig:umap_ISIC-2016} presents the two-dimensional (2D) UMAP embeddings of the ISIC-2016 dataset, where (a) contains 727 malignant and 173 benign lesions within the distribution, while (b), (c), and (d) in Fig. \ref{fig:umap_ISIC-2016} depict oversampled datasets featuring 2000 malignant and 2000 benign lesions each; it is evident from Fig. \ref{fig:umap_ISIC-2016} (b) that there is an overlap in the data distribution between the benign and malignant classes, whereas in Fig. \ref{fig:umap_ISIC-2016} (c) and Fig. \ref{fig:umap_ISIC-2016} (d), the distribution appears to separate into two distinct groups, highlighting the effectiveness of the FBGT and FAGT methods when compared to the No-Filtering approach.

Similarly, Fig. \ref{fig:umap_HAM10000} illustrates the three-dimensional (3D) UMAP embeddings of the HAM10000 dataset, where (a) shows a distribution comprising 9186 skin lesions, and (b), (c), and (d) in Fig. \ref{fig:umap_HAM10000} represent oversampled datasets, each displaying a subset of the 42,294 skin lesions; upon observing the 3D representations, it becomes evident that in Fig. \ref{fig:umap_HAM10000} (c) and Fig. \ref{fig:umap_HAM10000} (d), the data points for each class become more distinguishable within a three-dimensional space, whereas the unfiltered 3D representation in Fig. \ref{fig:umap_HAM10000} (b) demonstrates sparse data points, posing challenges for the classifier to identify specific regions for each class and consequently making the classification task more difficult.

The formation of clustered regions, as opposed to a single concentrated area, further substantiates the effectiveness of the FBGT and FAGT methods, addressing the issue of low inter-class variation within a dataset. By utilizing these techniques, we can substantially refine the classification process, ultimately leading to more accurate and dependable outcomes.

\section{Conclusion}
\label{sec:conclusion}
This paper introduces Cosine Similarity-based Image Filtering (CosSIF), a robust dataset filtering algorithm. We utilize CosSIF to create two filtering approaches: FBGT and FAGT. These methods rely on cosine similarity as the main metric for similarity calculation and aim to reduce the volume of GAN-generated synthetic images from the minority class that resemble similarity to images from the majority class. Our experimental results demonstrate that models trained on datasets processed with either the FBGT or FAGT methods show improved performance compared to models without these filtering methods. 
Through comprehensive experiments, we demonstrate that the proposed FAGT method, when applied to the ISIC-2016 dataset and trained on the ViT model, improves sensitivity by 1.59\% and AUC by 1.88\% compared to the baseline MelaNet. When we apply the FAGT and FBGT methods to the HAM10000 dataset and train them on the ConvNeXt and Swin Transformer models, we observe significant improvements in recall. Specifically, the FAGT method achieves a recall improvement of 13.72\% over the baseline IRv2+SA, with an accuracy of 94.44\%. Similarly, the FBGT method achieves a recall improvement of 13.75\% over the same baseline, with an accuracy of 94.04\%. For future research, our aim is to enhance the similarity calculation algorithm by incorporating a feature extraction and feature-based similarity calculation module. Moreover, we aim to apply our algorithm and filtering methods to various medical domains, including X-rays, CT scans, and MRI images. Furthermore, we plan to utilize the proposed CosSIF algorithm to develop a downsampling technique suitable for all image classification tasks.

\section*{Funding}
This research did not receive any specific grant from funding agencies in the public, commercial, or not-for-profit sectors.

\section*{CRediT authorship contribution statement}
\textbf{M. Islam:} Conceptualization, Methodology, Software, Validation, Formal analysis, Investigation, Resources, Data Curation, Writing - Original Draft, Visualization, Project administration.
\textbf{H. Zunair, N. Mohammed:} Resources, Investigation, Writing - Review \& Editing, Supervision, Funding acquisition.

\section*{Declaration of competing interest}
The authors declare that they have no known competing financial interests or personal relationships that could have appeared to
influence the work reported in this paper.

\section*{Data availability}
The datasets used in this research are publicly available and can be found at the following URLs: 
\url{https://challenge.isic-archive.com/data/} for the ISIC-2016 dataset and \url{https://dataverse.harvard.edu/dataset.xhtml?persistentId=doi:10.7910/DVN/DBW86T} for the HAM10000 dataset.

%% Loading bibliography style file
%\bibliographystyle{model1-num-names}
\bibliographystyle{cas-model2-names}

% Loading bibliography database
\bibliography{cas-refs}

\end{document}